\pdfoutput=1

\documentclass[11pt]{article}

\usepackage[preprint]{acl}

\usepackage{times}
\usepackage{latexsym}
\usepackage{hyperref}
\usepackage{csquotes}
\usepackage{color, colortbl}
\definecolor{Gray}{gray}{0.9}
\usepackage{array}
\usepackage{booktabs}
\usepackage{multirow}
\usepackage{multicol}
\usepackage{amsmath}
\usepackage{amssymb}
\usepackage{enumerate}
\usepackage{diagbox}
\usepackage{subcaption}
\usepackage{mathtools}
\usepackage{xcolor}
\usepackage{paralist}
\usepackage{lscape}
\usepackage{tabularx}
\usepackage{lipsum}
\usepackage{ragged2e}
\usepackage{makecell}
\usepackage{longtable}
\usepackage{comment}

\usepackage[T1]{fontenc}

\usepackage[utf8]{inputenc}

\usepackage{microtype}
\usepackage{soul}

\usepackage{inconsolata}

\usepackage{graphicx}

\newcommand{\dataset}{\textsc{arXiv2Table}}
\newcommand{\newmanDataset}{\textsc{ArxivDigesTables}}

\definecolor{colorxmark}{RGB}{255, 87, 51}
\definecolor{colorcmark}{RGB}{66, 154, 137}
\definecolor{headcolor}{HTML}{018161}
\definecolor{relationcolor}{HTML}{d95f02}
\definecolor{tailcolor}{HTML}{6560a3}

\title{\dataset: Toward Realistic Benchmarking and Evaluation for LLM-Based Literature-Review Table Generation}

\author{Weiqi Wang$^{\clubsuit\spadesuit}$,
Jiefu Ou$^{\clubsuit}$,
Yangqiu Song$^{\spadesuit}$,
Benjamin Van Durme$^{\clubsuit}$,
Daniel Khashabi$^{\clubsuit}$\\
$^{\clubsuit}$Center for Speech and Language Processing, Johns Hopkins University, Baltimore, MD, USA\\
$^{\spadesuit}$Department of Computer Science and Engineering, HKUST, Hong Kong SAR, China\\
\texttt{\{wwangbw, yqsong\}@cse.ust.hk, \{jou6, vandurme, danielk\}@jhu.edu}\\
}

\begin{document}
\maketitle
\begin{abstract}
Literature review tables are essential for summarizing and comparing collections of scientific papers.
In this paper, we study automatic generation of such tables from a pool of papers to satisfy a user's information need.
Building on recent work~\cite{ArxivDIGESTables}, we move beyond oracle settings by
(i) simulating \emph{well-specified yet schema-agnostic} user demands that avoid leaking gold column names or values,
(ii) explicitly modeling retrieval noise via semantically related but out-of-scope \emph{distractor} papers verified by human annotators,
and (iii) introducing a lightweight, annotation-free, utilization-oriented evaluation that decomposes utility (schema coverage, unary cell fidelity, pairwise relational consistency) and measures paper selection via a two-way QA procedure (gold$\rightarrow$system and system$\rightarrow$gold) with recall, precision, and F1.
To support reproducible evaluation, we introduce \dataset, a benchmark of 1,957 tables referencing 7,158 papers, with human-verified distractors and rewritten, schema-agnostic user demands.
We also develop an \emph{iterative, batch-based} generation method that co-refines paper filtering and schema over multiple rounds.
We validate the evaluation protocol with human audits and cross-evaluator checks.
Extensive experiments show that our method consistently improves over strong baselines, while absolute scores remain modest, underscoring the task's difficulty.
Our data and code is available at \href{https://github.com/JHU-CLSP/arXiv2Table}{https://github.com/JHU-CLSP/arXiv2Table}.
\end{abstract}

\section{Introduction}
Literature review tables play a crucial role in scientific research by organizing and summarizing large amounts of information from selected papers into a concise and comparable format~\cite{DBLP:conf/chi/RussellSPC93}.
At the core of these tables are the \textit{schema} and \textit{values} that define their structure, where \textit{schema} refers to the categories or aspects used to summarize different papers and \textit{values} correspond to the specific information extracted from each paper.
A well-defined \textit{schema} allows each work to be represented as a row of \textit{values}, enabling structured and transparent comparisons across different studies.

\begin{figure}[t]
     \centering
     \includegraphics[width=1\linewidth]{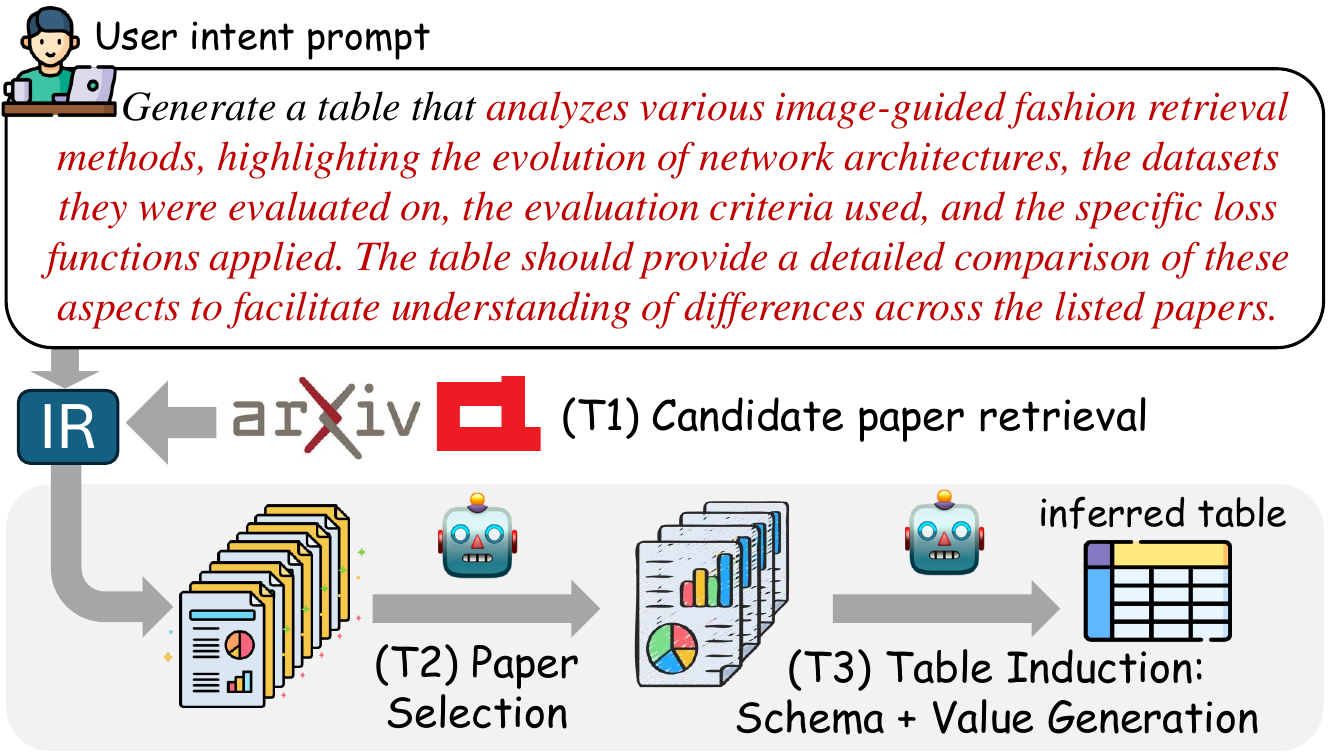}
     \caption{
     Overview of our proposed task: Given a user's demand, a simulated information retrieval (IR) engine first retrieves semantically relevant papers. 
     Then, a language model further filters them and induces the table's corresponding schema and values to satisfy the user's demand. 
     The \colorbox[HTML]{F2F2F2}{grayed region} indicates the scope covered by our method and benchmark (\dataset).
     }
    \label{fig:introduction_overview}
\end{figure}

With recent advancements in large language models (LLMs;~\citealp{o3,DeepSeekR1}), several studies~\cite{ArxivDIGESTables,dagdelen2024structured,DBLP:conf/aaai/SunHZMSC0024} have explored generating literature review tables by prompting LLMs with a set of pre-selected papers and the table's caption.
While these efforts represent meaningful progress, we argue that the existing task definition and evaluation protocols are somewhat unrealistic, thus hindering the practical applicability of generation methods.

First, existing pipelines assume that all provided papers are relevant and should be included in the table. 
However, in real-world scenarios, distractor papers—those that are irrelevant or contain limited useful information—are common~\cite{DeepSearch}. 
Models should be able to identify and filter out such papers before table construction.
Additionally, current pipelines use the ground-truth table's descriptive caption as the objective for generation.
These captions often lack sufficient context, making it difficult for LLMs to infer an appropriate schema, or they may inadvertently reveal the schema and values, leading to biased evaluations.

In this paper, we introduce our task, as illustrated in Figure~\ref{fig:introduction_overview}, which improves upon previous task definitions through two key adaptations.
First, our pilot study shows that LLMs struggle to retrieve relevant papers from large corpora.
To benchmark this, we introduce distractor papers by selecting them based on semantic similarity to papers in the ground-truth table.
LLMs must first determine which papers should be included before generating the table.
Second, we simulate \emph{well-specified yet schema-agnostic} user demands that describe the goal of curating the table without revealing column labels or cell values. 
This formulation is more realistic than terse captions while avoiding schema/value leakage that would bias evaluation.
We build upon the \newmanDataset~\cite{ArxivDIGESTables} dataset and construct a sibling benchmark through human annotation to verify the selected distractors, comprising 1,957 tables and 7,158 papers.
Meanwhile, current evaluation methods rely on static semantic embeddings to estimate schema overlap between generated and ground-truth tables and require human annotations to assess the quality of unseen schemas and values. 
However, semantic embeddings struggle to capture nuanced, context-specific variations due to their reliance on pre-trained representations, while human annotation is costly and time-consuming.
Moreover, the most effective table generation approaches define schemas primarily based on paper abstracts. 
This method risks missing important aspects present in the full text, leading to loosely defined schemas with inconsistent granularity.

To address these issues, we propose an annotation-free evaluation framework that instructs an LLM to synthesize QA pairs based on the ground-truth table and assess the generated table by answering these questions.
These QA pairs evaluate table content overlap across three dimensions: schema-level, single-cell, and pairwise-cell comparisons.
Additionally, we introduce a novel table generation method that batches input papers, iteratively refining paper selection and schema definition by revisiting each paper multiple times.
Extensive experiments using five LLMs demonstrate that they struggle with both selecting relevant papers and generating high-quality tables, while our method significantly improves performance on both fronts.
Expert validation further confirms the reliability of our QA-synthetic evaluations.

In summary, our contributions are threefold: 
(1) We introduce an improved task definition for literature review tabular generation, benchmarking it in a more realistic scenario by incorporating distractor papers and replacing table captions with abstract user demands;
(2) We propose an annotation-free evaluation framework that leverages LLM-generated QA pairs to assess schema-level, single-cell, and pairwise-cell content overlap, addressing the limitations of static semantic embeddings and human evaluation; and
(3) We develop a novel iterative batch-based table generation method that processes input papers in batches, refining schema definition and paper selection iteratively.

To the best of our knowledge, we are the first to introduce a task that simulates real-world use cases of scientific tabular generation by incorporating user demands and distractor papers, providing a more robust assessment of LLMs in this domain.

\section{Related Works}
\paragraph{Scientific literature tabular generation}
Prior works primarily attempt to generate scientific tables through two stages: schema induction and value extraction.
For schema induction, early methods like entity-based table generation~\cite{DBLP:conf/sigir/ZhangB18} focused on structured input, while recent work has explored schema induction from user queries~\cite{DBLP:journals/corr/abs-2404-13765} and comparative aspect extraction~\cite{DBLP:conf/sigir/HashimotoSYA17}.
For value extraction, various approaches such as document-grounded question-answering~\cite{DBLP:journals/tacl/KwiatkowskiPRCP19,DBLP:conf/naacl/DasigiLBCSG21,DBLP:conf/icml/LeeLPHKLL23}, aspect-based summarization~\cite{DBLP:conf/acl/AhujaXGHD22}, and document summarization~\cite{DBLP:conf/emnlp/DeYoungBZKW21,DBLP:conf/emnlp/LuDC20} have been proposed to extract relevant information.

Beyond these methods, several datasets have been introduced to support scientific table-related tasks, such as TableBank~\cite{DBLP:conf/lrec/LiCHWZL20}, SciGen~\cite{DBLP:conf/nips/MoosaviRRG21}, and SciTabQA~\cite{DBLP:conf/emnlp/LuPLNK23}.
\citet{DBLP:conf/emnlp/RamuGB24} propose an entailment-oriented evaluation complementary to our QA-based protocol.
Recently,~\citet{ArxivDIGESTables} proposed streamlining schema and value generation with LLMs sequentially and curated a large-scale benchmark for evaluation.
However, all these methods assume a clean and fully relevant set of papers and rely on predefined captions or abstract-based schemas
In contrast, we argue for an evaluation approach where candidate papers include tangentially relevant or distracting papers, aligning more closely with real-world literature review workflows~\cite{padmakumar2025setting}.

\paragraph{Table induction for general domains}
Other than the scientific domain, table induction is also widely studied as text-to-table generation. 
Prior works attempt this as a sequence-to-sequence task~\cite{DBLP:conf/acl/LiWSZWS23,DBLP:conf/acl/0001ZL22} or as a question-answering problem~\cite{DBLP:journals/corr/abs-2403-14457,DBLP:journals/corr/abs-2309-08963}.
Similar to these works, our framework is capable of better handling both structured and distractive input for real-world literature review and knowledge synthesis.

\section{Task Definition}
\label{sec:task_definition}
We first define a pipeline consisting of three subtasks that extend prior definitions and better capture the real-world usage of literature review tabular generation.
For all the following tasks, we are given a user demand prompt $p$, which specifies the intended purpose of creating the table.

\textbf{(T1) Candidate Paper Retrieval:}  
We begin with a given \textit{universe} of papers (e.g., the content of Google Scholar or arXiv) from which relevant papers need to be identified.
Given a large collection, the goal is to use a search engine (IR) to retrieve a subset of \textit{candidate} papers $C := \{d_i\}_{i=1}^{M}$ of size $M$, which may include distractor papers—i.e., papers that resemble the user demand prompt but do not fully satisfy the requirement.

\textbf{(T2) Paper Selection:}  
Given $C$, the second subtask is to select the \textit{relevant} subset of size $m$ ($m < M$): $R := \{d_i\}_{i=1}^{m} \subseteq C$, which best aligns with the user demand $p$.
T2 differs from T1 in scale. Due to the large scale of T1, IR engines must optimize for recall, ensuring that as many relevant papers as possible are retrieved. 
However, T2 operates at a smaller scale, where precision is the priority, as it focuses on filtering out distractors and selecting only the most relevant papers.

\textbf{(T3) Table Induction:}  
Given the selected papers $R$, the objective is to generate a table with $m$ rows and $N$ columns, where $N \geq 2$ (i.e., no single-column tables).
Each row $r_i \in \{r_1, r_2, \dots, r_m\}$ corresponds to a unique input document $d_i \in R$, and each column $c_j \in \{c_1, c_2, \dots, c_N\}$ represents a unique aspect of the documents.
We refer to these $N$ columns as the \textit{schema} of the table and the $N \times m$ cells as the \textit{values} of the table.
The value of each cell is derived from its respective document according to the aspect defined by the corresponding column.

\section{\dataset{} Construction}
\label{sec:dataset_construction}
We then construct~\dataset{} based on the~\newmanDataset{} dataset which consists of literature tables (extracted from computer science papers) and their corresponding captions. 
We filter out tables that are structurally incomplete or lack full text for all referenced papers.
As a result, we are left with 1,957 tables (with captions) which have rows referring to 7,158 papers. 
Our construction involves three pillars: user demand inference (\S\ref{sec:user_demand_inference}), a simulated paper retrieval (\S\ref{sec:difficulty_noisy_paper_retrieval}) and evaluation through utilization (\S\ref{sec:qa_synthesis_tabular_evaluation}). 

\subsection{Constructing User Demand Prompts}
\label{sec:user_demand_inference}
We simulate well-specified yet schema-agnostic user demands $p$ by rewriting original survey-table captions into user-demand-style prompts that better reflect how a researcher would request a comparison table.
Each rewritten prompt is required to be (i) \emph{self-contained}, meaning it is understandable without seeing the table, (ii) \emph{goal-oriented}, meaning it clearly states the purpose of the table, and (iii) \emph{schema/value non-leaking}, meaning it does not include gold column names, specific cell values, or direct paraphrases of them.

\paragraph{Table captions are not appropriate prompts}
While the input dataset contains one caption per table, collected from arXiv papers, these captions are meant to complement tables rather than fully describe them. As a result, they are generally concise.
For example, a table caption might read: \textit{``Performance comparison of different approaches,''} which is too vague to understand without seeing the table.
Consequently, using table captions as prompts may not yield a well-defined task.
A more contextually self-contained rewritten user demand might instead be: \textit{``Draft a table that compares different knowledge editing methods, focusing on their performance on QA datasets.''}
Operationally, a caption is deemed ``under-specified'' when it cannot unambiguously determine a comparison goal without seeing either the gold schema or table body.

\paragraph{Our prompt construction}
To address this issue, we rewrite the captions of literature review tables into abstract yet descriptive user intentions using GPT-4o.
We guide the model with a prompt (see \S\ref{appendix:implementation_details}) that explains the rewriting task and specifies that the resulting user demand should be sufficiently contextualized to clearly state the table's purpose while avoiding the inclusion or direct description of column names or specific values.
Here, the LLM is used solely for rewriting existing table captions into user demand prompts and for generating QA pairs grounded in ground-truth tables.
These reformulations are strictly tied to observed data and do not require external factual knowledge, minimizing risks of contamination or model-specific bias.

To enforce our constraints, we apply an automatic leak check immediately after rewriting (Appendix~\ref{appendix:implementation_details}) and discard or rewrite any demand that reveals schema or value information.
We also perform manual spot checks during construction to remove under-specified requests or prompts that implicitly reveal gold schema/value tokens.
For simplicity, we collect only one user demand per table.
More examples are provided in Appendix~\ref{appendix:case_studies}.

\begin{table}[t]
\footnotesize
\setlength{\tabcolsep}{3pt}
\centering
\resizebox{1\linewidth}{!}{%
\begin{tabular}{ll|cc}
\toprule
\textbf{Prompt} & \textbf{Content}   & \textbf{\#Table} $\downarrow$ & \textbf{\#Tokens} $\downarrow$ \\
\midrule
\multirow{2}{*}{Caption} & Schema  & 101 (5.2\%) & 1.2  \\
                         & Value   & 46 (2.4\%)  & 1.3 \\
\midrule
\multirow{2}{*}{User Demand} & Schema  & 14 (0.7\%) & 1.0 \\
                             & Value   & 8 (0.4\%)  & 1.0 \\
\bottomrule
\end{tabular}%
}
\caption{
Overlap statistics between prompts (the original caption or our constructed user demand) and table content (schema or values).
\textbf{\#Table:} Number (and \%) of tables with at least one token from table content overlapping with the prompt.
\textbf{\#Tokens:} Average count of overlapping tokens between table content and prompt.
}
\label{tab:schema_value_overlap}
\end{table}

\begin{figure}[t]
     \centering
     \includegraphics[width=1\linewidth]{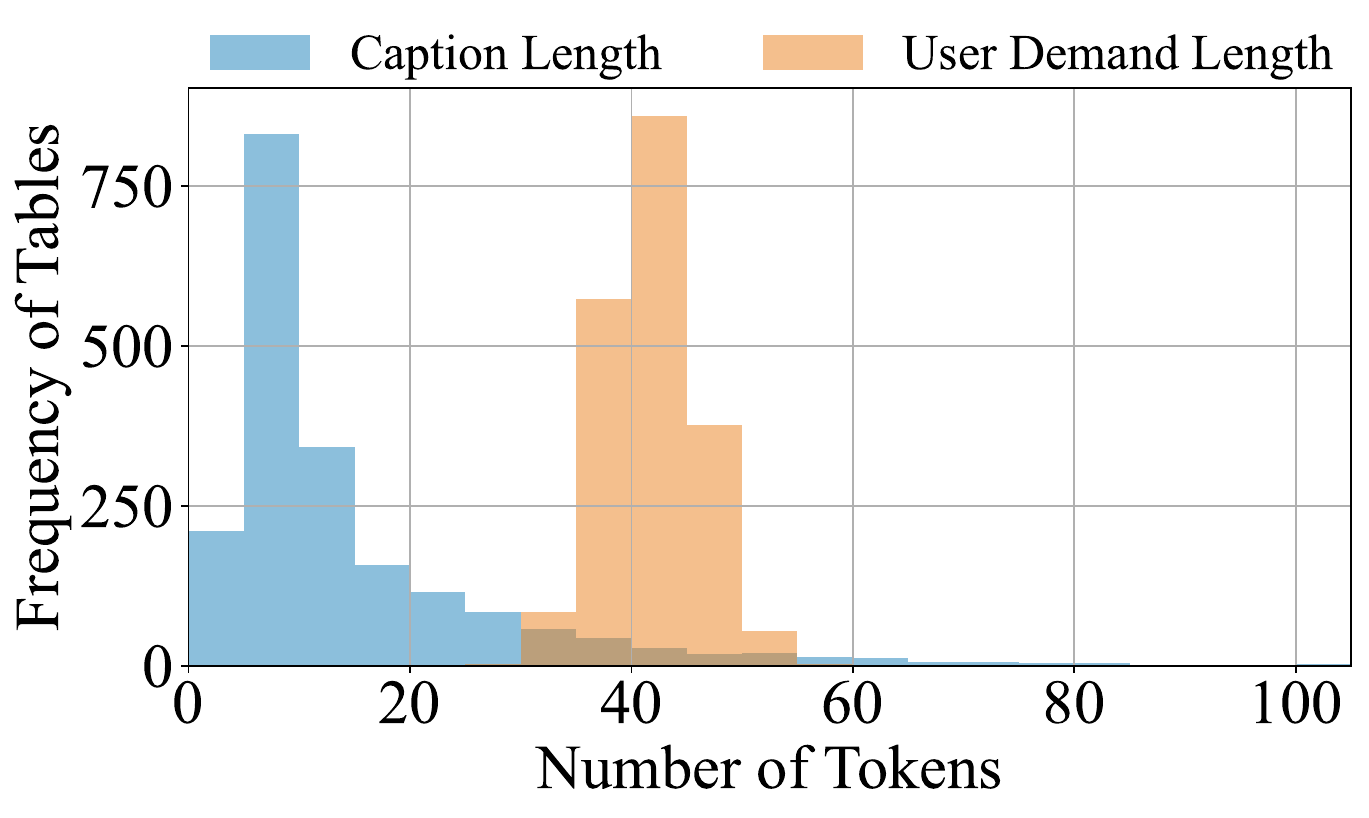}
     \caption{Distribution of the number of tokens between original captions and our modified user demands.
     }
    \label{fig:caption_user_demand_token_distribution}
\end{figure}

\paragraph{Table captions vs. constructed user demand prompts}
To verify that our collected user demands align with our objective, we visualize:
(1) the distribution of the number of tokens in the original and modified user demands, and
(2) the ratio of captions and user demands of different lengths that have token overlap with the schema or values.
From Figure~\ref{fig:caption_user_demand_token_distribution}, we observe that our modified user demands are generally longer than the original captions, providing a more detailed description of the table's goal.
Furthermore, as shown in Table~\ref{tab:schema_value_overlap}, user demands exhibit a significantly lower overlap ratio with the schema and table values, resulting in fewer overlapping tokens.
Quantitative statistics on lexical leakage and overlap are reported in the main paper.

\subsection{Paper Retrieval Simulation}
\label{sec:difficulty_noisy_paper_retrieval}
\begin{figure}[ht]
     \centering
     \includegraphics[width=1\linewidth]{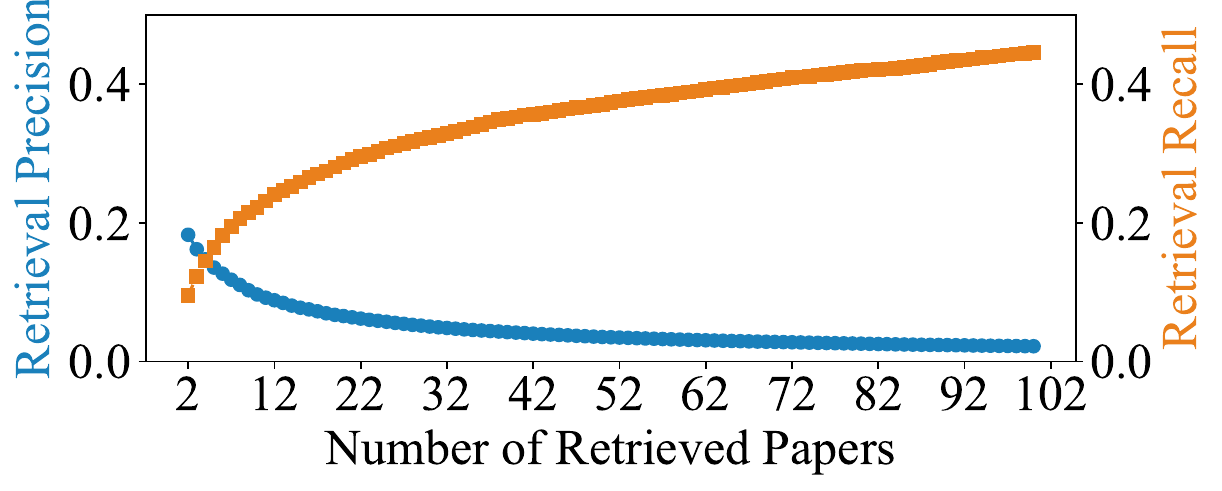}
     \caption{Precision and recall curves for different numbers of retrieved papers.}
     \label{fig:precision_recall}
\end{figure}

\paragraph{The unreliability of paper retrieval}
Next, we approach the first subtask, candidate paper retrieval, by conducting a pilot study to assess whether LMs can reliably retrieve relevant papers from a large corpus.  
For each table, we employ a SentenceBERT~\cite{DBLP:conf/emnlp/ReimersG19} encoder as a retrieval engine, selecting papers from the entire corpus based on the highest similarity between the table's user demand and each paper's title and abstract.  
We vary the number of retrieved papers between 2 and 100 and plot the precision and recall of retrieval against the ground-truth papers in the original table (Figure~\ref{fig:precision_recall}).  

We observe consistently low precision and recall across different retrieval sizes, highlighting the challenge of retrieving relevant papers from a noisy corpus.  
This demonstrates that the first subtask is non-trivial and may introduce noise into subtask T2.  
However, various information retrieval engines, such as Google Scholar and Semantic Scholar, can replace LMs in this subtask.  
Thus, we decide to simulate T1 by adding human-verified distractor papers into the candidate pool $C$, yielding a noisy input for T2 (paper selection on $C$ to produce $R$). 
This allows us to focus on evaluating LLMs' capabilities in the T2 and T3 subtasks.

\paragraph{Similarity-based paper retrieval}
Moving forward, we associate distractor paper candidates with each table to simulate a potentially noisy document pool before constructing the table.
Ideally, distractor candidates should be semantically related to the table but exhibit key differences that fail to meet the user demand.
To select such candidates, we adopt a retrieve-then-annotate approach. 
First, we use a SentenceBERT encoder $F$ to obtain embeddings for (1) the user demand $F(p)$ and (2) all papers in the corpus $\{F(d_i) \mid d_i \in U\}$. 
Each paper’s embedding is computed by encoding the concatenation of its title and abstract.
We then rank all papers $d_i \notin R$ based on the average of two cosine similarities: (1) the similarity between the candidate and the user demand, and (2) the average similarity between the candidate and each referenced paper:

{\small
$$s(d_i) =\cos(F(d_i), F(p)) + \frac{1}{m} \sum_{j=1}^{m} \cos(F(d_i), F(d_{u_j})).$$}

Higher values of $s(d_i)$ indicate stronger semantic relevance, and we select the top 10 ranked papers for each table as its distractor candidates.

\paragraph{Candidates verification via human annotation}
After selecting these candidates, we conduct human annotations to verify whether they should indeed be excluded from the table. 
Given that annotating these tables requires expert knowledge in computer science, we recruit a trained team of annotators with research experience in the field as annotators.
To ensure they are well-prepared for the task, the annotators undergo rigorous training, including pilot annotation exams. 
Their task is to make a binary decision on whether a given distractor paper—based on its title, abstract, user demand, the ground-truth table, and the titles and abstracts of all referenced papers—should be included in the table.

Each table contains annotations for 10 papers, with each distractor paper initially assigned to two randomly selected annotators. 
If both annotators agree on the label, it is finalized.
Otherwise, two additional annotators review the paper until a consensus is reached.
In the first round, the inter-annotator agreement (IAA) is 94\% based on pairwise agreement, and the Fleiss’ Kappa~\cite{fleiss1971measuring} score is 0.73, indicating a substantial level of agreement.
Finally, for each table, we randomly select a number of distractor papers between $[m,10]$ and merge them with $R$ to form $C$.

\begin{figure}[t]
     \centering
     \includegraphics[width=1\linewidth,trim=10.9cm 0.7cm 5.1cm 0.8cm,clip=true]{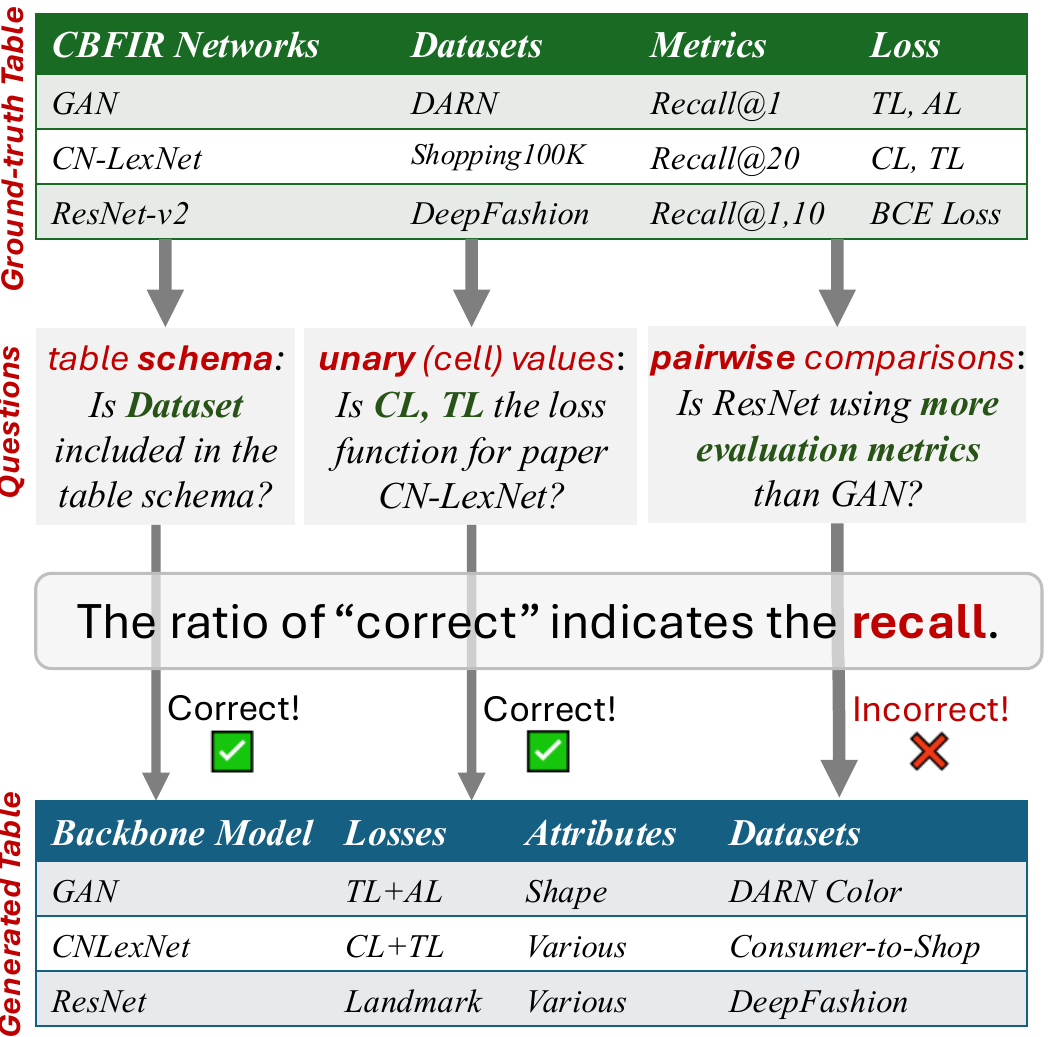}
     \caption{Overview of our proposed LLM-based QA-synthesis evaluation protocol, where LLMs synthesize QA pairs based on the ground-truth table and utilize the generated table to answer them.
     The ratio of successfully answered QA pairs indicate the ratio of information preserved.}
     \label{fig:method_overview}
\end{figure}

\subsection{Evaluation via LLM-based Utilization}
\label{sec:qa_synthesis_tabular_evaluation}
After constructing the benchmark, we propose evaluating the quality of generated tables from a utilization perspective to address the challenge of aligning schemas and values despite potential differences in phrasing.  
This is achieved by synthesizing QA pairs based on the ground-truth table and using the generated table to answer them, or vice versa.  
The flexibility of this QA synthesis allows us to evaluate multiple dimensions of the table while ensuring a structured and scalable assessment.  
An overview with running examples is shown in Figure~\ref{fig:method_overview}.  

\paragraph{Dimensions of evaluating a table with QAs}  
We introduce three key aspects for evaluating a table in terms of its usability:  
(1) \textbf{Schema}: whether a specific column is included in the generated schema,  
(2) \textbf{Unary Value}: whether a particular cell from the ground-truth table appears in the generated table,
(3) \textbf{Pairwise Value}: whether relationships between two cells remain consistent in the generated table.  

\paragraph{Recall evaluation}  
We guide GPT-4o in generating binary QA pairs based on the ground-truth table.
For the first two aspects, we generate QA pairs for all columns and cells, whereas for the third, we randomly sample 10 cell pairs per table and synthesize them into QA pairs.
We then prompt GPT-4o to answer these questions based on the generated table, providing yes/no responses.
If the answer cannot be found, the model is instructed to respond with ``no'', and vice versa for ``yes''.
The ratio of ``yes'' answers indicates how well the generated table preserves the schema, individual values, and pairwise relationships.
This represents the \textbf{recall} of the ground-truth table, measuring how much original information is retained in generations.

\paragraph{Precision evaluation}  
To additionally evaluate \textbf{precision}, we reverse the process: instead of generating QA pairs from the ground-truth table, we generate them from the generated table and ask another LLM (by default, we use the same backbone for consistency; cross-model swaps are reported in Appendix~\ref{app:evaluator_checks}) to answer them using the ground-truth table.  
The precision score reflects how much of the generated table's content is actually supported by the original data.  
By computing the ratio of ``yes'' answers, precision quantifies how much of the generated table is supported by the ground-truth table; novel content beyond gold is not credited under this automatic protocol.
To assess evaluator dependence, we swap QA synthesizer/answerer model families and observe stable method rankings; results are in Appendix~\ref{app:evaluator_checks}.

\section{Tabular Generation Methodologies}
We explore a range of methods to evaluate on our proposed task, starting from several baselines inspired by prior work (\S\ref{sec:baseline_methods}) and then our proposed approach (\S\ref{sec:batch_tabular_generation_method}). 

\subsection{Baseline Methods}
\label{sec:baseline_methods}
We first introduce three methods for generating literature review tables to evaluate their performance on our task and use them as baselines for our proposed method.
For easy reference, these methods are termed numerically.

First, \textbf{Baseline 1} generates the table in a one-step process.
It takes all available papers $R$ and the user demand $p$ as input, and the model is asked to select all relevant papers and output a table with a well-defined schema and filled values in a single round of conversation.
However, this method struggles with extremely long prompts that exceed the LLMs' context window when generating large tables.

To address this issue, \textbf{Baseline 2} processes papers individually.
For each document, the model decides whether it should be included based on the user demand.
If included, the model generates a table for that document.
After processing all documents, the final table is created by merging the schemas of all individual tables using exact string matching and copying the corresponding values.
While this approach reduces the input prompt length, it results in highly sparse tables due to inconsistent schema across papers and the potential omission of relevant information when individual papers lack sufficient context to define comprehensive table aspects.

To overcome both issues,~\citet{ArxivDIGESTables} introduce a two-stage process.
In the first stage, the model selects papers relevant to the user demand based on their titles and abstracts, then generates a corresponding schema.
In the second stage, the model loops through the selected papers and fills in the respective rows based on the full text of each document.
A minor drawback of this method is that the schema is generated solely from titles and abstracts, which may overlook details present only in the full text.
Note that this method is the \textbf{strongest recent baseline} for scientific tabular generation while other text-to-table methods~\cite{T3} are not directly applicable due to different assumptions.

To probe whether explicit reasoning mitigates multi-step synthesis errors, we also evaluate COT-augmented variants that add lightweight chain-of-thought style scaffolding to the strongest baseline and to our method.

\begin{table*}[t]
    \small
    \centering
    \resizebox{1\linewidth}{!}{
    \begin{tabular}{@{}l|l|ccc|ccc|ccc|ccc|c@{}}
    \toprule
    \multirow{2}{*}{\textbf{Backbone Model}} & \multirow{2}{*}{\textbf{Method}} 
    & \multicolumn{3}{c|}{\textbf{Paper (T2)}} 
    & \multicolumn{3}{c|}{\textbf{Schema}} 
    & \multicolumn{3}{c|}{\textbf{Unary Value}} 
    & \multicolumn{3}{c|}{\textbf{Pairwise Value}} 
    & \multirow{2}{*}{\textbf{Avg}} \\
    \cmidrule(lr){3-5} \cmidrule(lr){6-8} \cmidrule(lr){9-11} \cmidrule(lr){12-14}
    & & \textbf{R} & \textbf{P} & \textbf{F1} 
      & \textbf{P} & \textbf{R} & \textbf{F1} 
      & \textbf{P} & \textbf{R} & \textbf{F1} 
      & \textbf{P} & \textbf{R} & \textbf{F1} \\
    \midrule
    \multirow{4}{*}{\textbf{LLaMa-3.3 (70B)}} 
      & Baseline 1        & 52.8 & 50.0 & 51.4 & 31.3 & 37.7 & 34.2 & 29.6 & 40.4 & 34.2 & 28.4 & 31.8 & 30.0 & 32.8\\
      & Baseline 2        & 65.4 & 63.0 & 64.2 & 26.7 & \textbf{\underline{69.3}} & 38.5 & 17.0 & 56.8 & 26.2 & 11.2 & 22.5 & 15.0 & 26.6\\
      & Newman et al.     & 61.9 & 60.0 & 60.9 & 36.4 & 40.5 & 38.3 & 32.8 & 44.5 & 37.8 & 29.5 & 30.2 & 29.8 & 35.3\\
      & \textbf{Ours}     & \underline{69.3} & \underline{66.5} & \underline{67.9} & \underline{41.9} & 55.4 & \underline{47.7} & \underline{43.1} & \underline{62.6} & \underline{51.1} & \underline{36.4} & \underline{46.9} & \underline{41.0} & \underline{46.6}\\
    \midrule
    \multirow{4}{*}{\textbf{Mistral-Large (123B)}} 
      & Baseline 1        & 54.7 & 51.5 & 53.0 & 33.1 & 34.5 & 33.8 & 31.6 & 30.4 & 31.0 & 15.5 & 24.7 & 19.0 & 27.9\\
      & Baseline 2        & 66.8 & 64.0 & 65.4 & 27.4 & \underline{65.0} & 38.5 & 22.7 & 47.4 & 30.7 & 17.8 & 30.7 & 22.6 & 30.6\\
      & Newman et al.     & 67.9 & 65.5 & 66.7 & 39.9 & 41.6 & 40.7 & 34.7 & 46.3 & 39.7 & 29.9 & 35.1 & 32.3 & 37.6\\
      & \textbf{Ours}     & \underline{71.3} & \underline{68.0} & \underline{69.6} & \underline{45.4} & 56.7 & \underline{50.4} & \underline{43.3} & \underline{61.5} & \underline{50.8} & \underline{42.0} & \underline{49.2} & \underline{45.3} & \underline{48.8}\\
    \midrule
    \multirow{4}{*}{\textbf{DeepSeek-V3 (685B)}} 
      & Baseline 1        & 57.5 & 55.0 & 56.2 & 38.7 & 41.7 & 40.1 & 32.5 & 43.8 & 37.3 & 28.7 & 31.8 & 30.1 & 35.8\\
      & Baseline 2        & 69.8 & 67.0 & 68.4 & 34.9 & \underline{69.0} & 46.4 & 27.1 & 55.5 & 36.4 & 25.7 & 32.7 & 28.8 & 37.2\\
      & Newman et al.     & 70.9 & 68.5 & 69.7 & 39.4 & 44.2 & 41.7 & 36.6 & 49.2 & 42.0 & 33.3 & 36.5 & 34.8 & 39.5\\
      & \textbf{Ours}     & \underline{74.3} & \underline{71.0} & \underline{72.6} & \underline{39.6} & 56.9 & \underline{46.7} & \underline{47.7} & \underline{65.2} & \underline{55.1} & \underline{40.4} & \underline{49.8} & \underline{44.6} & \underline{48.8}\\
    \midrule
    \multirow{4}{*}{\textbf{GPT-4o-mini}} 
      & Baseline 1        & 55.9 & 53.0 & 54.4 & 32.0 & 35.7 & 33.7 & 28.9 & 39.3 & 33.3 & 25.0 & 31.0 & 27.7 & 31.6\\
      & Baseline 2        & 68.2 & 65.0 & 66.6 & 31.5 & \underline{67.7} & 43.0 & 27.7 & 50.8 & 35.9 & 21.6 & 28.3 & 24.5 & 34.5\\
      & Newman et al.     & 69.3 & 66.0 & 67.6 & 40.3 & 45.9 & 42.9 & 38.3 & 47.5 & 42.4 & 35.0 & 37.8 & 36.3 & 40.5\\
      & \textbf{Ours}     & \underline{72.6} & \underline{69.0} & \underline{70.8} & \underline{46.5} & 59.7 & \underline{52.3} & \textbf{\underline{49.0}} & \textbf{\underline{66.7}} & \textbf{\underline{56.5}} & \underline{43.5} & \underline{51.9} & \underline{47.3} & \underline{52.0}\\
    \midrule
    \multirow{6}{*}{\textbf{GPT-4o}} 
      & Baseline 1        & 58.5 & 56.0 & 57.2 & 35.8 & 43.2 & 39.2 & 36.9 & 41.8 & 39.2 & 29.0 & 34.7 & 31.6 & 36.7\\
      & Baseline 2        & 70.2 & 67.0 & 68.6 & 34.2 & \underline{68.0} & 45.5 & 27.9 & 56.0 & 37.2 & 19.4 & 33.6 & 24.6 & 35.8\\
      & Newman et al.     & 71.3 & 68.5 & 69.9 & 45.0 & 47.9 & 46.4 & 38.7 & 49.8 & 43.6 & 36.9 & 40.0 & 38.4 & 42.8\\
      & Newman et al. + COT & 72.5 & \multicolumn{1}{c}{--} & \multicolumn{1}{c|}{--} & 47.0 & 49.0 & 48.0 & 40.0 & 51.0 & 45.0 & 39.0 & 42.0 & 40.5 & 44.5\\
      & \textbf{Ours}     & \textbf{\underline{74.6}} & \textbf{\underline{71.5}} & \textbf{\underline{73.0}} & 51.5 & 59.4 & 55.2 & 46.1 & 66.7 & 54.5 & 45.9 & 55.7 & 50.3 & 53.3\\
      & \textbf{Ours + COT} & \underline{75.8} & \multicolumn{1}{c}{--} & \multicolumn{1}{c|}{--} & \underline{53.0} & \underline{60.5} & \underline{56.5} & \underline{48.0} & \underline{68.0} & \underline{56.3} & \underline{47.0} & \underline{57.0} & \underline{51.5} & \underline{55.0}\\
    \midrule
    \multirow{1}{*}{\textbf{GPT-o3}} 
      & \textbf{Ours}     & \textbf{\underline{77.2}} & \multicolumn{1}{c}{--} & \multicolumn{1}{c|}{--} & \textbf{55.0} & \textbf{62.0} & \textbf{58.3} & \textbf{50.0} & \textbf{70.0} & \textbf{58.3} & \textbf{49.0} & \textbf{59.0} & \textbf{53.5} & \textbf{56.7}\\
    \bottomrule
    \end{tabular}
    }
\caption{Tabular evaluation results (\%) on \dataset{}. All tasks use P/R/F1 as evaluation metrics. The best within each \emph{backbone} is \underline{underlined} and the best \emph{overall} is \textbf{bold}.
}
\vspace{-0.1in}
\label{tab:maintable_eval_results}
\end{table*}

\subsection{Iterative Batch-based Tabular Generation}
\label{sec:batch_tabular_generation_method}
Then, we introduce our proposed method for generating literature review tables.
Our approach consists of three steps: (A) key information extraction, (B) paper batching, and (C) paper selection and schema refinement, where the latter two steps can be iterated multiple times.

\paragraph{(A) Key Information Extraction}  
Processing multiple papers simultaneously using their full text often results in excessively long prompts that exceed the LLMs' context window.  
To address this, we first shorten each paper by instructing the LLM to extract key information from the full text that is relevant to the user's requirements.  
Notably, we do not rely solely on the abstract, as important details often appear in the full text but are omitted from the abstract.  
For each paper, we provide the LLM with its title, abstract, and full text, along with the user’s request, and ask it to generate a concise paragraph that preserves all potentially relevant details.  
These summary paragraphs serve as condensed representations of the papers for subsequent processing.  

\paragraph{(B) Paper Batching}  
Next, we divide all key information paragraphs into smaller batches.  
Processing too many papers at once negatively affects the model's performance (as demonstrated by the comparison of Baseline 1 in Table~\ref{tab:maintable_eval_results}), whereas batching facilitates more efficient comparisons within each batch.  
For simplicity, we set a batch size of 4 and randomly partition $R$ into $\left\lceil \frac{|R|}{4} \right\rceil$ batches.

\paragraph{(C) Paper Selection and Schema Refinement}  
We initialize an empty schema and table, then sequentially process each batch with the LLM by providing it with the user’s request and summaries of batched papers.  
The LLM is instructed to (1) decide whether each paper should be included or removed based on its key information and (2) refine the schema based on the current batch of papers. 
Schema refinement involves adding or removing specific columns or modifying existing values to align with different formats.  
For new papers that are deemed suitable for inclusion yet are not in the current table, we prompt the LLM to insert a new row with the extracted fields.
This ensures that the table remains dynamically structured, continuously adapting to new information while maintaining consistency across batches.

Afterward, we iterate steps B and C for $k$ iterations.
Here $k$ is a hyper-parameter and we set $k=5$ in our experiments.  
The rationale is that multiple iterations allow the schema and table contents to progressively improve, ensuring better alignment with user demands.  
In each iteration, the batches are newly randomized so that each paper is compared with different subsets, enabling more robust decision-making and reducing bias from specific batch compositions.  
This iterative refinement also mitigates errors from earlier batches by revisiting and adjusting prior decisions based on newly processed information.  
After the final iteration, we re-verify each populated cell directly against the source text; unsupported values are set to \textsc{N/A}.

\begin{figure*}[t]
     \centering
     \includegraphics[width=1\linewidth]{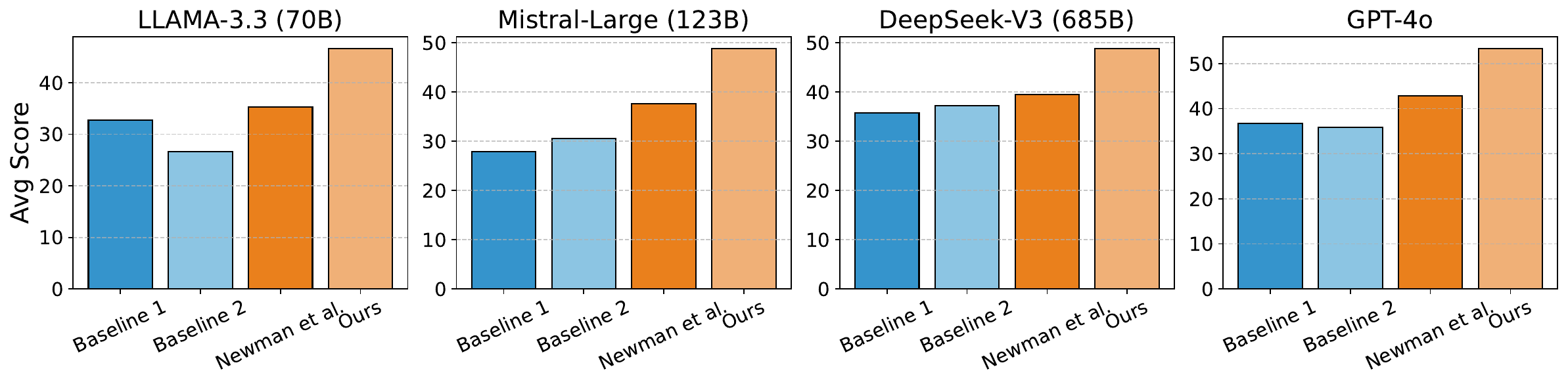}
     \caption{Average performance scores of four backbone LLMs across four different methods. The comparison highlights the consistent improvement of our proposed method over existing baselines and prior work.}
     \vspace{-0.1in}
    \label{fig:results_overview}
\end{figure*}

\section{Experiments and Analyses}
\subsection{Experiment Setup}
To demonstrate the generalizability of our method and evaluations, we conduct experiments using three proprietary and three open-source LLMs as backbone model representatives: GPT-4o~\cite{GPT4o}, GPT-4o-mini~\cite{GPT4omini}, GPT-o3~\cite{GPTo3}, DeepSeek-V3 (685B;~\citealp{DeepSeek-V3}), LLaMa-3.3 (70B;~\citealp{LLaMa3}), and Mistral-Large (123B;~\citealp{MistralLarge}).
We apply all baseline methods and our proposed method to each model and use our evaluation framework to assess the quality of the generated tables based on our benchmark, focusing on four aspects: paper selection (\textbf{Paper}), schema content overlap (\textbf{Schema}), single-cell value overlap (\textbf{Unary Value}), and comparisons across cells (\textbf{Pairwise Value})\footnote{For paper selection, recall is emphasized instead of precision since missing relevant papers is more critical than including slightly noisy ones in literature review tasks.}.
For paper selection (T2) and all T3 dimensions, we report precision (P), recall (R), and F1. 
For the COT and GPT-o3 variants, T2 P/F1 are omitted on this slice due to compute limits.
Due to cost, GPT-o3 is evaluated on a fixed subset.

\subsection{Main Evaluation Results}
We report the main results in Table~\ref{tab:maintable_eval_results} and summarize key findings below; a model-wise comparison across methods is shown in Figure~\ref{fig:results_overview}.

\noindent\textbf{(1) All methods and models struggle to distinguish relevant papers from distractors.}
Even at their best settings, LLaMa-3.3 and GPT-4o reach only 65.4--74.6\% recall and 60.0--71.5\% precision in T2 (Table~\ref{tab:maintable_eval_results}), indicating substantial distractor inclusion or missed relevant papers depending on the operating point.
We also find that processing papers individually, or using only abstracts for inclusion decisions, performs better than concatenating full texts, suggesting that overly long prompts can weaken per-paper inclusion decisions.

\noindent\textbf{(2) Aligning generated schemas with the ground-truth table remains challenging.}
Among the baselines, the second method tends to achieve higher recall (e.g., 69.3\% with LLaMa-3.3), largely because it produces more columns and thus overlaps more often with the ground-truth schema.
Other methods have notably lower recall, indicating that generating meaningful columns that match the gold structure remains difficult.

\noindent\textbf{(3) While unary values are well preserved, pairwise comparisons suffer substantial losses.}
Most methods, especially ours, achieve relatively strong unary F1 scores, whereas extracting and preserving pairwise relationships remains challenging.
This trend holds across models: systems often identify individual entries correctly but fail to capture relations between them, highlighting the difficulty of preserving complex relational comparisons in generated tables.

\noindent\textbf{(4) Our proposed method improves performance across all aspects and models.}
Across backbone models and evaluation criteria, our method consistently outperforms the baselines.
It achieves the strongest overall results and the highest unary/pairwise F1, demonstrating robustness in both distractor handling and precise table generation.

\noindent\textbf{(5) Larger models lead to better performance.}
For open-source LLMs, increasing model size consistently improves results under the same method.
For GPT-4o, adding CoT yields consistent T3 gains with similar or slightly higher token budgets, while GPT-o3 attains the highest T3 scores overall, suggesting that stronger backbones better exploit iterative batching.

\subsection{Validation of Utilization-Based Evaluation}
To verify the reliability of synthesizing QA pairs using LLMs for evaluating tabular data, we conduct two complementary expert assessments. 
First, we invited the authors (as domain experts) to manually inspect a random sample of 200 QA pairs—spanning schema-level, unary value, and pairwise value comparisons. 
Annotators were asked to assess (1) whether each QA pair is firmly grounded in the source table, and (2) whether the LLM's answer is correct based on the generated target table. 
As shown in Table~\ref{tab:expert_validation}, the expert acceptance rates exceed 98\% in all categories, confirming the quality of the synthesized QA pairs.
Note that Table~\ref{tab:human_alignment} measures evaluator–human agreement (reliability), not end-to-end table quality, so inter-method gaps are expected to be smaller than in Table~\ref{tab:maintable_eval_results}.

\begin{table}[t]
\small
\centering
\resizebox{1\linewidth}{!}{
\begin{tabular}{l|ccc}
    \toprule
   Table & Schema & Unary Value & Pairwise Value \\
    \midrule
    Source & 99.5\% & 100\% & 98.5\% \\
    Target & 98.5\% & 99.5\% & 97.0\% \\
    \bottomrule
\end{tabular}
}
\caption{Expert acceptance rate for the synthesized QA pairs sampled from our evaluations.}
\label{tab:expert_validation}
\end{table}

Second, we conducted an additional human study to assess whether our LLM-based evaluation aligns with human judgment across different generation methods. 
For each method, we sampled 300 QA pairs, answered them using both LLMs and human annotators, and measured the agreement rate. 
As shown in Table~\ref{tab:human_alignment}, LLM and human “yes” response rates are highly consistent, with over 97\% agreement across all methods. 
These results reinforce the robustness of our evaluation framework, demonstrating that LLM-synthesized QA pairs provide a scalable and trustworthy proxy for human judgment in assessing semantically diverse tabular outputs.
Specifically, these results indicate that the high agreement is not driven by an inherent bias of LLMs toward their own generated QA pairs.

\begin{table}[t]
\small
\centering
\resizebox{1\linewidth}{!}{
\begin{tabular}{@{}l|ccc@{}}
    \toprule
    Method & LLM Rate & Human Rate & Agreement \\
    \midrule
    Baseline 1 & 39.1\% & 39.6\% & 97.3\% \\
    Baseline 2 & 57.1\% & 57.3\% & 98.2\% \\
    Newman et al. & 42.9\% & 43.0\% & 98.6\% \\
    Ours     & 57.3\% & 57.5\% & 98.0\% \\
    \bottomrule
\end{tabular}
}
\caption{Comparison between GPT-4o and human annotators on 300 QA pairs. We report the proportion of ``yes'' answers by each and their overall agreement.}
\vspace{-0.1in}
\label{tab:human_alignment}
\end{table}

\subsection{Batch Size and Iteration Sensitivity}
\label{app:batch_sensitivity}
We analyze the effect of batch size $b$ and iteration count $k$ in our iterative pipeline, which repeatedly performs paper selection and schema/table refinement over multiple batches.
Using GPT-4o as the backbone, we evaluate 60 tables sampled to cover a range of sizes and schema complexity.
We vary $b\in\{2,4,6\}$ and $k\in\{2,\dots,5\}$, and report T3 macro-F1, the average of Schema, Unary, and Pairwise F1.
Each configuration is run with two seeds and averaged, with standard deviation shown where relevant.
Token budgets follow the main-experiment prompt templates, and per-iteration context lengths remain below 128K.

Table~\ref{tab:bk_grid} shows that performance improves steadily over the first few iterations, confirming the benefit of iterative refinement across different paper subsets.
Gains saturate around $k\approx4$--5, with $b{=}4$ performing slightly better overall.
The best macro-F1 is achieved at $b{=}4,k{=}4$ (53.1), while $k{=}5$ yields almost no further improvement.
Across seeds, variability is modest ($\leq 0.5$ points).

\begin{table}[h]
\small\centering
\begin{tabular}{@{}c|ccc|c@{}}
\toprule
$k$ & \multicolumn{3}{c|}{Macro-F1 by batch size} & Std.\ dev.\ (avg) \\
& $b{=}2$ & $b{=}4$ & $b{=}6$ & over seeds \\
\midrule
2 & 50.1 & 50.7 & 50.5 & 0.5 \\
3 & 51.8 & 52.2 & 52.0 & 0.4 \\
4 & 52.6 & \textbf{53.1} & 52.9 & 0.4 \\
5 & \textbf{52.7} & 53.0 & \textbf{52.9} & 0.4 \\
\bottomrule
\end{tabular}
\caption{Macro-F1 across iterations $k$ and batch sizes $b$. Gains saturate by $k\approx4$--5, with $b{=}4$ slightly ahead.}
\label{tab:bk_grid}
\end{table}

\begin{figure}[t]
     \centering
     \includegraphics[width=0.9\linewidth]{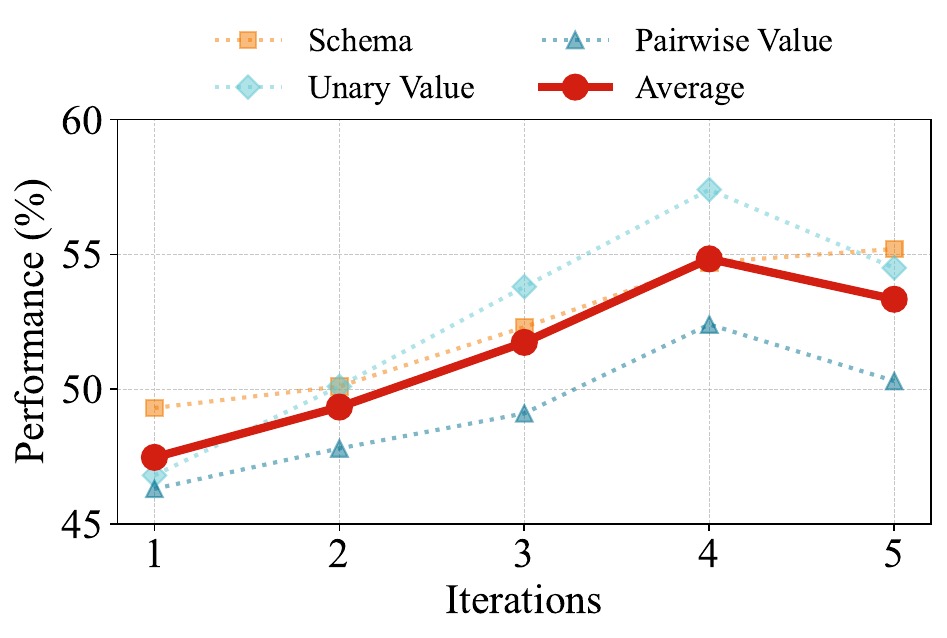}
     \caption{Ablation study on the number of iterations for our iterative batch-based table generation method.}
     \vspace{-0.1in}
    \label{fig:ablation_study_iteration}
\end{figure}

Figure~\ref{fig:ablation_study_iteration} shows the same trend across iterations.
Most gains occur before iteration 4, after which performance plateaus.
Later rounds may also introduce unsupported values that slightly reduce precision.
We further observe that pairwise F1 benefits the most, suggesting that repeated cross-batch comparisons mainly improve relational consistency.
Overall, these results support $k{=}4$ or $k{=}5$ as a practical default.

\section{Conclusions}
In this work, we introduce an improved literature review table generation task that incorporates distractor papers and replaces table captions with abstract user demands to better align with real-world scenarios, and curated an associated benchmark.
Additionally, we propose an annotation-free evaluation framework using LLM-synthesized QA pairs and a novel method to enhance table generation.
Our experiments show that current LLMs and existing methods struggle with our task, while our approach significantly improves performance.
We envision that our work paves the way for more automated and scalable literature review table generation, ultimately facilitating the efficient synthesis of scientific knowledge in large-scale applications.

\newpage
\section*{Limitations}
A minor limitation is that our work uses~\newmanDataset{} as the source of literature review tables for subsequent data reconstruction. 
However, \citet{ArxivDIGESTables} have included their pipeline for scalably extracting literature review tables from scientific papers, thus resolving the data reliance gap~\cite{DBLP:conf/emnlp/OuWSJSCJWLMHWMRKD25,DBLP:journals/corr/abs-2504-13834}.
Beyond the computer science domain, our formulation and methodology are readily applicable to other scientific fields such as medicine, physics, and social sciences, where structured comparisons across publications are equally valuable. 
Moreover, the core task—generating structured tables from noisy, unstructured input with under-specified intent—extends naturally to real-world applications like news fact aggregation, personalized knowledge card generation, and structured database population from web or legal documents.

Another limitation of our work is its reliance on GPT-4o, a proprietary LLM, for benchmark curation and subsequent evaluation, which may introduce several issues.
First, it raises concerns about data contamination~\cite{DBLP:conf/naacl/Deng0TGC24,DBLP:conf/acl/DongJLJGYL24}, as the model may generate user demands (during benchmark curation) and synthesis evaluation questions (when evaluating a generated table against the ground truth) that are similar to its training data, potentially leading to inflated performance in table generation.
A data provenance check~\cite{DBLP:conf/icml/LongpreMOBSGPK24} can be further implemented to address this issue.
Second, the benchmark and evaluation process may inherit the internal knowledge or semantic distribution biases of GPT-4o, which could skew the evaluation of other LLMs and reduce the generalizability of our findings. 
Lastly, a minor issue is scalability, as curating larger datasets using a proprietary model can be resource-intensive and may limit accessibility when extending our framework to other literature or domains.
Future work can explore the use of open-source LLMs to replicate the entire process for convenient adaptation to other tabular datasets.

\section*{Ethics Statement}
The \newmanDataset{}~\cite{ArxivDIGESTables} dataset used in our work is shared under the Open Data Commons License, which grants us access to it and allows us to improve and redistribute it for research purposes.
Regarding language models, we access all open-source LMs via the Hugging Face Hub~\cite{DBLP:conf/emnlp/WolfDSCDMCRLFDS20} and proprietary GPT models through their official API\footnote{\href{https://platform.openai.com/}{https://platform.openai.com/}}.
The number of these models, if available, is marked in Table~\ref{tab:maintable_eval_results}.
All associated licenses for these models permit user access for research purposes, and we commit to following all terms of use.

When prompting GPT-4o to generate user demands and synthetic QA questions, we explicitly state in the prompt that the LLM should not generate any content that contains personal privacy violations, promotes violence, racial discrimination, hate speech, sexual, or self-harm contents.
We also manually inspect a random sample of 100 data entries generated by GPT-4o for offensive content, and none are detected. 
Therefore, we believe that our dataset is safe and will not yield any negative or harmful impact.

Our human annotations are conducted by trained graduate-level annotators ($n{=}19$), compensated above local minimum wage.
They have sufficient experience in data collection for training large language models and are proficient in English, primarily from Asia. 
They receive thorough training on the task and are reminded to have a clear understanding of the task instructions before proceeding to annotation.
The high level of inter-agreement also confirms the quality of our annotation.
The expert annotators have agreed to participate as their contribution to the paper without receiving any compensation.

\section*{Acknowledgments}
We thank Muhan Gao and the JHU CLSP community for their discussions and inspiration, and the HKUST KnowComp community for their help with data annotation.
The authors of this paper were supported by the ITSP Platform Research Project (ITS/189/23FP) from the Innovation and Technology Commission of Hong Kong SAR, China; the AoE (AoE/E-601/24-N), RIF (R6021-20), and GRF (16205322) from the Research Grants Council of Hong Kong SAR, China; and the ONR grant (N0001424-1-2089) from the U.S. Office of Naval Research.

\bibliography{custom}

@inproceedings{ArxivDIGESTables,
  author       = {Benjamin Newman and
                  Yoonjoo Lee and
                  Aakanksha Naik and
                  Pao Siangliulue and
                  Raymond Fok and
                  Juho Kim and
                  Daniel S. Weld and
                  Joseph Chee Chang and
                  Kyle Lo},
  editor       = {Yaser Al{-}Onaizan and
                  Mohit Bansal and
                  Yun{-}Nung Chen},
  title        = {ArxivDIGESTables: Synthesizing Scientific Literature into Tables using
                  Language Models},
  booktitle    = {Proceedings of the 2024 Conference on Empirical Methods in Natural
                  Language Processing, {EMNLP} 2024, Miami, FL, USA, November 12-16,
                  2024},
  pages        = {9612--9631},
  publisher    = {Association for Computational Linguistics},
  year         = {2024},
  url          = {https://aclanthology.org/2024.emnlp-main.538},
  bibsource    = {dblp computer science bibliography, https://dblp.org}
}

@inproceedings{T3,
  author       = {Zheye Deng and
                  Chunkit Chan and
                  Weiqi Wang and
                  Yuxi Sun and
                  Wei Fan and
                  Tianshi Zheng and
                  Yauwai Yim and
                  Yangqiu Song},
  editor       = {Yaser Al{-}Onaizan and
                  Mohit Bansal and
                  Yun{-}Nung Chen},
  title        = {Text-Tuple-Table: Towards Information Integration in Text-to-Table
                  Generation via Global Tuple Extraction},
  booktitle    = {Proceedings of the 2024 Conference on Empirical Methods in Natural
                  Language Processing, {EMNLP} 2024, Miami, FL, USA, November 12-16,
                  2024},
  pages        = {9300--9322},
  publisher    = {Association for Computational Linguistics},
  year         = {2024},
  url          = {https://aclanthology.org/2024.emnlp-main.523},
  bibsource    = {dblp computer science bibliography, https://dblp.org}
}

@article{GPT4o,
  title={Hello GPT-4o},
  author={OpenAI},
  journal={OpenAI},
  year={2024},
  url={https://openai.com/index/hello-gpt-4o/}
}

@article{GPT4omini,
  title={GPT-4o mini: advancing cost-efficient intelligence},
  author={OpenAI},
  journal={OpenAI},
  year={2024},
  url={https://openai.com/index/gpt-4o-mini-advancing-cost-efficient-intelligence/}
}

@article{GPTo3,
  title={Introducing OpenAI o3 and o4-mini},
  author={OpenAI},
  journal={OpenAI},
  year={2025},
  url={https://openai.com/index/introducing-o3-and-o4-mini/}
}

@article{DeepSeek-V3,
  author       = {DeepSeek{-}AI and
                  Aixin Liu and
                  Bei Feng and
                  Bing Xue and
                  Bingxuan Wang and
                  Bochao Wu and
                  Chengda Lu and
                  Chenggang Zhao and
                  Chengqi Deng and
                  Chenyu Zhang and
                  Chong Ruan and
                  Damai Dai and
                  Daya Guo and
                  Dejian Yang and
                  Deli Chen and
                  Dongjie Ji and
                  Erhang Li and
                  Fangyun Lin and
                  Fucong Dai and
                  Fuli Luo and
                  Guangbo Hao and
                  Guanting Chen and
                  Guowei Li and
                  H. Zhang and
                  Han Bao and
                  Hanwei Xu and
                  Haocheng Wang and
                  Haowei Zhang and
                  Honghui Ding and
                  Huajian Xin and
                  Huazuo Gao and
                  Hui Li and
                  Hui Qu and
                  J. L. Cai and
                  Jian Liang and
                  Jianzhong Guo and
                  Jiaqi Ni and
                  Jiashi Li and
                  Jiawei Wang and
                  Jin Chen and
                  Jingchang Chen and
                  Jingyang Yuan and
                  Junjie Qiu and
                  Junlong Li and
                  Junxiao Song and
                  Kai Dong and
                  Kai Hu and
                  Kaige Gao and
                  Kang Guan and
                  Kexin Huang and
                  Kuai Yu and
                  Lean Wang and
                  Lecong Zhang and
                  Lei Xu and
                  Leyi Xia and
                  Liang Zhao and
                  Litong Wang and
                  Liyue Zhang and
                  Meng Li and
                  Miaojun Wang and
                  Mingchuan Zhang and
                  Minghua Zhang and
                  Minghui Tang and
                  Mingming Li and
                  Ning Tian and
                  Panpan Huang and
                  Peiyi Wang and
                  Peng Zhang and
                  Qiancheng Wang and
                  Qihao Zhu and
                  Qinyu Chen and
                  Qiushi Du and
                  R. J. Chen and
                  R. L. Jin and
                  Ruiqi Ge and
                  Ruisong Zhang and
                  Ruizhe Pan and
                  Runji Wang and
                  Runxin Xu and
                  Ruoyu Zhang and
                  Ruyi Chen and
                  S. S. Li and
                  Shanghao Lu and
                  Shangyan Zhou and
                  Shanhuang Chen and
                  Shaoqing Wu and
                  Shengfeng Ye and
                  Shengfeng Ye and
                  Shirong Ma and
                  Shiyu Wang and
                  Shuang Zhou and
                  Shuiping Yu and
                  Shunfeng Zhou and
                  Shuting Pan and
                  T. Wang and
                  Tao Yun and
                  Tian Pei and
                  Tianyu Sun and
                  W. L. Xiao and
                  Wangding Zeng},
  title        = {DeepSeek-V3 Technical Report},
  journal      = {CoRR},
  volume       = {abs/2412.19437},
  year         = {2024},
  url          = {https://doi.org/10.48550/arXiv.2412.19437},
  doi          = {10.48550/ARXIV.2412.19437},
  eprinttype    = {arXiv},
  eprint       = {2412.19437},
  bibsource    = {dblp computer science bibliography, https://dblp.org}
}

@misc{DeepSeekR1,
      title={DeepSeek-R1: Incentivizing Reasoning Capability in LLMs via Reinforcement Learning}, 
      author={DeepSeek-AI and Daya Guo and Dejian Yang and Haowei Zhang and Junxiao Song and Ruoyu Zhang and Runxin Xu and Qihao Zhu and Shirong Ma and Peiyi Wang and Xiao Bi and Xiaokang Zhang and Xingkai Yu and Yu Wu and Z. F. Wu and Zhibin Gou and Zhihong Shao and Zhuoshu Li and Ziyi Gao and Aixin Liu and Bing Xue and Bingxuan Wang and Bochao Wu and Bei Feng and Chengda Lu and Chenggang Zhao and Chengqi Deng and Chenyu Zhang and Chong Ruan and Damai Dai and Deli Chen and Dongjie Ji and Erhang Li and Fangyun Lin and Fucong Dai and Fuli Luo and Guangbo Hao and Guanting Chen and Guowei Li and H. Zhang and Han Bao and Hanwei Xu and Haocheng Wang and Honghui Ding and Huajian Xin and Huazuo Gao and Hui Qu and Hui Li and Jianzhong Guo and Jiashi Li and Jiawei Wang and Jingchang Chen and Jingyang Yuan and Junjie Qiu and Junlong Li and J. L. Cai and Jiaqi Ni and Jian Liang and Jin Chen and Kai Dong and Kai Hu and Kaige Gao and Kang Guan and Kexin Huang and Kuai Yu and Lean Wang and Lecong Zhang and Liang Zhao and Litong Wang and Liyue Zhang and Lei Xu and Leyi Xia and Mingchuan Zhang and Minghua Zhang and Minghui Tang and Meng Li and Miaojun Wang and Mingming Li and Ning Tian and Panpan Huang and Peng Zhang and Qiancheng Wang and Qinyu Chen and Qiushi Du and Ruiqi Ge and Ruisong Zhang and Ruizhe Pan and Runji Wang and R. J. Chen and R. L. Jin and Ruyi Chen and Shanghao Lu and Shangyan Zhou and Shanhuang Chen and Shengfeng Ye and Shiyu Wang and Shuiping Yu and Shunfeng Zhou and Shuting Pan and S. S. Li and Shuang Zhou and Shaoqing Wu and Shengfeng Ye and Tao Yun and Tian Pei and Tianyu Sun and T. Wang and Wangding Zeng and Wanjia Zhao and Wen Liu and Wenfeng Liang and Wenjun Gao and Wenqin Yu and Wentao Zhang and W. L. Xiao and Wei An and Xiaodong Liu and Xiaohan Wang and Xiaokang Chen and Xiaotao Nie and Xin Cheng and Xin Liu and Xin Xie and Xingchao Liu and Xinyu Yang and Xinyuan Li and Xuecheng Su and Xuheng Lin and X. Q. Li and Xiangyue Jin and Xiaojin Shen and Xiaosha Chen and Xiaowen Sun and Xiaoxiang Wang and Xinnan Song and Xinyi Zhou and Xianzu Wang and Xinxia Shan and Y. K. Li and Y. Q. Wang and Y. X. Wei and Yang Zhang and Yanhong Xu and Yao Li and Yao Zhao and Yaofeng Sun and Yaohui Wang and Yi Yu and Yichao Zhang and Yifan Shi and Yiliang Xiong and Ying He and Yishi Piao and Yisong Wang and Yixuan Tan and Yiyang Ma and Yiyuan Liu and Yongqiang Guo and Yuan Ou and Yuduan Wang and Yue Gong and Yuheng Zou and Yujia He and Yunfan Xiong and Yuxiang Luo and Yuxiang You and Yuxuan Liu and Yuyang Zhou and Y. X. Zhu and Yanhong Xu and Yanping Huang and Yaohui Li and Yi Zheng and Yuchen Zhu and Yunxian Ma and Ying Tang and Yukun Zha and Yuting Yan and Z. Z. Ren and Zehui Ren and Zhangli Sha and Zhe Fu and Zhean Xu and Zhenda Xie and Zhengyan Zhang and Zhewen Hao and Zhicheng Ma and Zhigang Yan and Zhiyu Wu and Zihui Gu and Zijia Zhu and Zijun Liu and Zilin Li and Ziwei Xie and Ziyang Song and Zizheng Pan and Zhen Huang and Zhipeng Xu and Zhongyu Zhang and Zhen Zhang},
      year={2025},
      eprint={2501.12948},
      archivePrefix={arXiv},
      primaryClass={cs.CL},
      url={https://arxiv.org/abs/2501.12948}, 
}

@inproceedings{padmakumar2025setting,
  title={Setting The Table with Intent: Intent-aware Schema Generation and Editing for Literature Review Tables},
  author={Padmakumar, Vishakh and Chang, Joseph Chee and Lo, Kyle and Downey, Doug and Naik, Aakanksha},
  year={2025},
  note={To appear},
  url={https://github.com/vishakhpk/arxivdigestables-with-intent}
}

@article{LLAMA3,
  author       = {Abhimanyu Dubey and
                  Abhinav Jauhri and
                  Abhinav Pandey and
                  Abhishek Kadian and
                  Ahmad Al{-}Dahle and
                  Aiesha Letman and
                  Akhil Mathur and
                  Alan Schelten and
                  Amy Yang and
                  Angela Fan and
                  Anirudh Goyal and
                  Anthony Hartshorn and
                  Aobo Yang and
                  Archi Mitra and
                  Archie Sravankumar and
                  Artem Korenev and
                  Arthur Hinsvark and
                  Arun Rao and
                  Aston Zhang and
                  Aur{\'{e}}lien Rodriguez and
                  Austen Gregerson and
                  Ava Spataru and
                  Baptiste Rozi{\`{e}}re and
                  Bethany Biron and
                  Binh Tang and
                  Bobbie Chern and
                  Charlotte Caucheteux and
                  Chaya Nayak and
                  Chloe Bi and
                  Chris Marra and
                  Chris McConnell and
                  Christian Keller and
                  Christophe Touret and
                  Chunyang Wu and
                  Corinne Wong and
                  Cristian Canton Ferrer and
                  Cyrus Nikolaidis and
                  Damien Allonsius and
                  Daniel Song and
                  Danielle Pintz and
                  Danny Livshits and
                  David Esiobu and
                  Dhruv Choudhary and
                  Dhruv Mahajan and
                  Diego Garcia{-}Olano and
                  Diego Perino and
                  Dieuwke Hupkes and
                  Egor Lakomkin and
                  Ehab AlBadawy and
                  Elina Lobanova and
                  Emily Dinan and
                  Eric Michael Smith and
                  Filip Radenovic and
                  Frank Zhang and
                  Gabriel Synnaeve and
                  Gabrielle Lee and
                  Georgia Lewis Anderson and
                  Graeme Nail and
                  Gr{\'{e}}goire Mialon and
                  Guan Pang and
                  Guillem Cucurell and
                  Hailey Nguyen and
                  Hannah Korevaar and
                  Hu Xu and
                  Hugo Touvron and
                  Iliyan Zarov and
                  Imanol Arrieta Ibarra and
                  Isabel M. Kloumann and
                  Ishan Misra and
                  Ivan Evtimov and
                  Jade Copet and
                  Jaewon Lee and
                  Jan Geffert and
                  Jana Vranes and
                  Jason Park and
                  Jay Mahadeokar and
                  Jeet Shah and
                  Jelmer van der Linde and
                  Jennifer Billock and
                  Jenny Hong and
                  Jenya Lee and
                  Jeremy Fu and
                  Jianfeng Chi and
                  Jianyu Huang and
                  Jiawen Liu and
                  Jie Wang and
                  Jiecao Yu and
                  Joanna Bitton and
                  Joe Spisak and
                  Jongsoo Park and
                  Joseph Rocca and
                  Joshua Johnstun and
                  Joshua Saxe and
                  Junteng Jia and
                  Kalyan Vasuden Alwala and
                  Kartikeya Upasani and
                  Kate Plawiak and
                  Ke Li and
                  Kenneth Heafield and
                  Kevin Stone and
                  et al.},
  title        = {The Llama 3 Herd of Models},
  journal      = {CoRR},
  volume       = {abs/2407.21783},
  year         = {2024},
  url          = {https://doi.org/10.48550/arXiv.2407.21783},
  doi          = {10.48550/ARXIV.2407.21783},
  eprinttype    = {arXiv},
  eprint       = {2407.21783},
  bibsource    = {dblp computer science bibliography, https://dblp.org}
}

@inproceedings{DBLP:conf/emnlp/RamuGB24,
  author       = {Pritika Ramu and
                  Aparna Garimella and
                  Sambaran Bandyopadhyay},
  editor       = {Yaser Al{-}Onaizan and
                  Mohit Bansal and
                  Yun{-}Nung Chen},
  title        = {Is This a Bad Table? {A} Closer Look at the Evaluation of Table Generation
                  from Text},
  booktitle    = {Proceedings of the 2024 Conference on Empirical Methods in Natural
                  Language Processing, {EMNLP} 2024, Miami, FL, USA, November 12-16,
                  2024},
  pages        = {22206--22216},
  publisher    = {Association for Computational Linguistics},
  year         = {2024},
  url          = {https://doi.org/10.18653/v1/2024.emnlp-main.1239},
  doi          = {10.18653/V1/2024.EMNLP-MAIN.1239},
  bibsource    = {dblp computer science bibliography, https://dblp.org}
}

@article{MistralLarge,
  title={Large Enough},
  author={Mistral-AI},
  journal={Mistral AI Blog},
  year={2024},
  url={https://mistral.ai/news/mistral-large-2407/}
}

@inproceedings{DBLP:conf/naacl/Deng0TGC24,
  author       = {Chunyuan Deng and
                  Yilun Zhao and
                  Xiangru Tang and
                  Mark Gerstein and
                  Arman Cohan},
  editor       = {Kevin Duh and
                  Helena G{\'{o}}mez{-}Adorno and
                  Steven Bethard},
  title        = {Investigating Data Contamination in Modern Benchmarks for Large Language
                  Models},
  booktitle    = {Proceedings of the 2024 Conference of the North American Chapter of
                  the Association for Computational Linguistics: Human Language Technologies
                  (Volume 1: Long Papers), {NAACL} 2024, Mexico City, Mexico, June 16-21,
                  2024},
  pages        = {8706--8719},
  publisher    = {Association for Computational Linguistics},
  year         = {2024},
  url          = {https://doi.org/10.18653/v1/2024.naacl-long.482},
  doi          = {10.18653/V1/2024.NAACL-LONG.482},
  bibsource    = {dblp computer science bibliography, https://dblp.org}
}

@inproceedings{DBLP:conf/icml/LongpreMOBSGPK24,
  author       = {Shayne Longpre and
                  Robert Mahari and
                  Naana Obeng{-}Marnu and
                  William Brannon and
                  Tobin South and
                  Katy Ilonka Gero and
                  Alex Pentland and
                  Jad Kabbara},
  title        = {Position: Data Authenticity, Consent, {\&} Provenance for {AI}
                  are all broken: what will it take to fix them?},
  booktitle    = {Forty-first International Conference on Machine Learning, {ICML} 2024,
                  Vienna, Austria, July 21-27, 2024},
  publisher    = {OpenReview.net},
  year         = {2024},
  url          = {https://openreview.net/forum?id=3hSTecKy1b},
  bibsource    = {dblp computer science bibliography, https://dblp.org}
}

@inproceedings{DBLP:conf/acl/DongJLJGYL24,
  author       = {Yihong Dong and
                  Xue Jiang and
                  Huanyu Liu and
                  Zhi Jin and
                  Bin Gu and
                  Mengfei Yang and
                  Ge Li},
  editor       = {Lun{-}Wei Ku and
                  Andre Martins and
                  Vivek Srikumar},
  title        = {Generalization or Memorization: Data Contamination and Trustworthy
                  Evaluation for Large Language Models},
  booktitle    = {Findings of the Association for Computational Linguistics, {ACL} 2024,
                  Bangkok, Thailand and virtual meeting, August 11-16, 2024},
  pages        = {12039--12050},
  publisher    = {Association for Computational Linguistics},
  year         = {2024},
  url          = {https://doi.org/10.18653/v1/2024.findings-acl.716},
  doi          = {10.18653/V1/2024.FINDINGS-ACL.716},
  bibsource    = {dblp computer science bibliography, https://dblp.org}
}

@article{o3,
  title={OpenAI o3-mini},
  author={OpenAI},
  journal={OpenAI Blog},
  year={2025},
  url={https://openai.com/index/openai-o3-mini/}
}

@article{dagdelen2024structured,
  title={Structured information extraction from scientific text with large language models},
  author={Dagdelen, John and Dunn, Alexander and Lee, Sanghoon and Walker, Nicholas and Rosen, Andrew S and Ceder, Gerbrand and Persson, Kristin A and Jain, Anubhav},
  journal={Nature Communications},
  volume={15},
  number={1},
  pages={1418},
  year={2024},
  publisher={Nature Publishing Group UK London}
}

@inproceedings{DBLP:conf/aaai/SunHZMSC0024,
  author       = {Liangtai Sun and
                  Yang Han and
                  Zihan Zhao and
                  Da Ma and
                  Zhennan Shen and
                  Baocai Chen and
                  Lu Chen and
                  Kai Yu},
  editor       = {Michael J. Wooldridge and
                  Jennifer G. Dy and
                  Sriraam Natarajan},
  title        = {SciEval: {A} Multi-Level Large Language Model Evaluation Benchmark
                  for Scientific Research},
  booktitle    = {Thirty-Eighth {AAAI} Conference on Artificial Intelligence, {AAAI}
                  2024, Thirty-Sixth Conference on Innovative Applications of Artificial
                  Intelligence, {IAAI} 2024, Fourteenth Symposium on Educational Advances
                  in Artificial Intelligence, {EAAI} 2014, February 20-27, 2024, Vancouver,
                  Canada},
  pages        = {19053--19061},
  publisher    = {{AAAI} Press},
  year         = {2024},
  url          = {https://doi.org/10.1609/aaai.v38i17.29872},
  doi          = {10.1609/AAAI.V38I17.29872},
  bibsource    = {dblp computer science bibliography, https://dblp.org}
}

@article{DeepSearch,
  title={Introducing deep research},
  author={OpenAI},
  journal={OpenAI Blog},
  year={2025},
  url={https://openai.com/index/introducing-deep-research/}
}

@inproceedings{DBLP:conf/chi/RussellSPC93,
  author       = {Daniel M. Russell and
                  Mark Stefik and
                  Peter Pirolli and
                  Stuart K. Card},
  editor       = {Bert Arnold and
                  Gerrit C. van der Veer and
                  Ted N. White},
  title        = {The cost structure of sensemaking},
  booktitle    = {Human-Computer Interaction, {INTERACT} '93, {IFIP} {TC13} International
                  Conference on Human-Computer Interaction, 24-29 April 1993, Amsterdam,
                  The Netherlands, jointly organised with {ACM} Conference on Human
                  Aspects in Computing Systems CHI'93},
  pages        = {269--276},
  publisher    = {{ACM}},
  year         = {1993},
  url          = {https://doi.org/10.1145/169059.169209},
  doi          = {10.1145/169059.169209},
  bibsource    = {dblp computer science bibliography, https://dblp.org}
}

@inproceedings{DBLP:conf/sigir/ZhangB18,
  author       = {Shuo Zhang and
                  Krisztian Balog},
  editor       = {Kevyn Collins{-}Thompson and
                  Qiaozhu Mei and
                  Brian D. Davison and
                  Yiqun Liu and
                  Emine Yilmaz},
  title        = {On-the-fly Table Generation},
  booktitle    = {The 41st International {ACM} {SIGIR} Conference on Research {\&}
                  Development in Information Retrieval, {SIGIR} 2018, Ann Arbor, MI,
                  USA, July 08-12, 2018},
  pages        = {595--604},
  publisher    = {{ACM}},
  year         = {2018},
  url          = {https://doi.org/10.1145/3209978.3209988},
  doi          = {10.1145/3209978.3209988},
  bibsource    = {dblp computer science bibliography, https://dblp.org}
}

@article{DBLP:journals/corr/abs-2404-13765,
  author       = {Xingbo Wang and
                  Samantha L. Huey and
                  Rui Sheng and
                  Saurabh Mehta and
                  Fei Wang},
  title        = {SciDaSynth: Interactive Structured Knowledge Extraction and Synthesis
                  from Scientific Literature with Large Language Model},
  journal      = {CoRR},
  volume       = {abs/2404.13765},
  year         = {2024},
  url          = {https://doi.org/10.48550/arXiv.2404.13765},
  doi          = {10.48550/ARXIV.2404.13765},
  eprinttype    = {arXiv},
  eprint       = {2404.13765},
  bibsource    = {dblp computer science bibliography, https://dblp.org}
}

@inproceedings{DBLP:conf/sigir/HashimotoSYA17,
  author       = {Hayato Hashimoto and
                  Kazutoshi Shinoda and
                  Hikaru Yokono and
                  Akiko Aizawa},
  editor       = {Philipp Mayr and
                  Muthu Kumar Chandrasekaran and
                  Kokil Jaidka},
  title        = {Automatic Generation of Review Matrices as Multi-document Summarization
                  of Scientific Papers},
  booktitle    = {Proceedings of the 2nd Joint Workshop on Bibliometric-enhanced Information
                  Retrieval and Natural Language Processing for Digital Libraries {(BIRNDL}
                  2017) co-located with the 40th International {ACM} {SIGIR} Conference
                  on Research and Development in Information Retrieval {(SIGIR} 2017),
                  Tokyo, Japan, August 11, 2017},
  series       = {{CEUR} Workshop Proceedings},
  volume       = {1888},
  pages        = {69--82},
  publisher    = {CEUR-WS.org},
  year         = {2017},
  url          = {https://ceur-ws.org/Vol-1888/paper6.pdf},
  bibsource    = {dblp computer science bibliography, https://dblp.org}
}

@article{DBLP:journals/tacl/KwiatkowskiPRCP19,
  author       = {Tom Kwiatkowski and
                  Jennimaria Palomaki and
                  Olivia Redfield and
                  Michael Collins and
                  Ankur P. Parikh and
                  Chris Alberti and
                  Danielle Epstein and
                  Illia Polosukhin and
                  Jacob Devlin and
                  Kenton Lee and
                  Kristina Toutanova and
                  Llion Jones and
                  Matthew Kelcey and
                  Ming{-}Wei Chang and
                  Andrew M. Dai and
                  Jakob Uszkoreit and
                  Quoc Le and
                  Slav Petrov},
  title        = {Natural Questions: a Benchmark for Question Answering Research},
  journal      = {Trans. Assoc. Comput. Linguistics},
  volume       = {7},
  pages        = {452--466},
  year         = {2019},
  url          = {https://doi.org/10.1162/tacl\_a\_00276},
  doi          = {10.1162/TACL\_A\_00276},
  bibsource    = {dblp computer science bibliography, https://dblp.org}
}

@inproceedings{DBLP:conf/naacl/DasigiLBCSG21,
  author       = {Pradeep Dasigi and
                  Kyle Lo and
                  Iz Beltagy and
                  Arman Cohan and
                  Noah A. Smith and
                  Matt Gardner},
  editor       = {Kristina Toutanova and
                  Anna Rumshisky and
                  Luke Zettlemoyer and
                  Dilek Hakkani{-}T{\"{u}}r and
                  Iz Beltagy and
                  Steven Bethard and
                  Ryan Cotterell and
                  Tanmoy Chakraborty and
                  Yichao Zhou},
  title        = {A Dataset of Information-Seeking Questions and Answers Anchored in
                  Research Papers},
  booktitle    = {Proceedings of the 2021 Conference of the North American Chapter of
                  the Association for Computational Linguistics: Human Language Technologies,
                  {NAACL-HLT} 2021, Online, June 6-11, 2021},
  pages        = {4599--4610},
  publisher    = {Association for Computational Linguistics},
  year         = {2021},
  url          = {https://doi.org/10.18653/v1/2021.naacl-main.365},
  doi          = {10.18653/V1/2021.NAACL-MAIN.365},
  bibsource    = {dblp computer science bibliography, https://dblp.org}
}

@inproceedings{DBLP:conf/icml/LeeLPHKLL23,
  author       = {Yoonjoo Lee and
                  Kyungjae Lee and
                  Sunghyun Park and
                  Dasol Hwang and
                  Jaehyeon Kim and
                  Hong{-}In Lee and
                  Moontae Lee},
  editor       = {Andreas Krause and
                  Emma Brunskill and
                  Kyunghyun Cho and
                  Barbara Engelhardt and
                  Sivan Sabato and
                  Jonathan Scarlett},
  title        = {{QASA:} Advanced Question Answering on Scientific Articles},
  booktitle    = {International Conference on Machine Learning, {ICML} 2023, 23-29 July
                  2023, Honolulu, Hawaii, {USA}},
  series       = {Proceedings of Machine Learning Research},
  volume       = {202},
  pages        = {19036--19052},
  publisher    = {{PMLR}},
  year         = {2023},
  url          = {https://proceedings.mlr.press/v202/lee23n.html},
  bibsource    = {dblp computer science bibliography, https://dblp.org}
}

@inproceedings{DBLP:conf/acl/AhujaXGHD22,
  author       = {Ojas Ahuja and
                  Jiacheng Xu and
                  Akshay Gupta and
                  Kevin Horecka and
                  Greg Durrett},
  editor       = {Smaranda Muresan and
                  Preslav Nakov and
                  Aline Villavicencio},
  title        = {{ASPECTNEWS:} Aspect-Oriented Summarization of News Documents},
  booktitle    = {Proceedings of the 60th Annual Meeting of the Association for Computational
                  Linguistics (Volume 1: Long Papers), {ACL} 2022, Dublin, Ireland,
                  May 22-27, 2022},
  pages        = {6494--6506},
  publisher    = {Association for Computational Linguistics},
  year         = {2022},
  url          = {https://doi.org/10.18653/v1/2022.acl-long.449},
  doi          = {10.18653/V1/2022.ACL-LONG.449},
  bibsource    = {dblp computer science bibliography, https://dblp.org}
}

@inproceedings{DBLP:conf/emnlp/DeYoungBZKW21,
  author       = {Jay DeYoung and
                  Iz Beltagy and
                  Madeleine van Zuylen and
                  Bailey Kuehl and
                  Lucy Lu Wang},
  editor       = {Marie{-}Francine Moens and
                  Xuanjing Huang and
                  Lucia Specia and
                  Scott Wen{-}tau Yih},
  title        = {MS{\textbackslash}{\^{}}2: Multi-Document Summarization of Medical
                  Studies},
  booktitle    = {Proceedings of the 2021 Conference on Empirical Methods in Natural
                  Language Processing, {EMNLP} 2021, Virtual Event / Punta Cana, Dominican
                  Republic, 7-11 November, 2021},
  pages        = {7494--7513},
  publisher    = {Association for Computational Linguistics},
  year         = {2021},
  url          = {https://doi.org/10.18653/v1/2021.emnlp-main.594},
  doi          = {10.18653/V1/2021.EMNLP-MAIN.594},
  bibsource    = {dblp computer science bibliography, https://dblp.org}
}

@inproceedings{DBLP:conf/emnlp/LuDC20,
  author       = {Yao Lu and
                  Yue Dong and
                  Laurent Charlin},
  editor       = {Bonnie Webber and
                  Trevor Cohn and
                  Yulan He and
                  Yang Liu},
  title        = {Multi-XScience: {A} Large-scale Dataset for Extreme Multi-document
                  Summarization of Scientific Articles},
  booktitle    = {Proceedings of the 2020 Conference on Empirical Methods in Natural
                  Language Processing, {EMNLP} 2020, Online, November 16-20, 2020},
  pages        = {8068--8074},
  publisher    = {Association for Computational Linguistics},
  year         = {2020},
  url          = {https://doi.org/10.18653/v1/2020.emnlp-main.648},
  doi          = {10.18653/V1/2020.EMNLP-MAIN.648},
  bibsource    = {dblp computer science bibliography, https://dblp.org}
}

@inproceedings{DBLP:conf/lrec/LiCHWZL20,
  author       = {Minghao Li and
                  Lei Cui and
                  Shaohan Huang and
                  Furu Wei and
                  Ming Zhou and
                  Zhoujun Li},
  editor       = {Nicoletta Calzolari and
                  Fr{\'{e}}d{\'{e}}ric B{\'{e}}chet and
                  Philippe Blache and
                  Khalid Choukri and
                  Christopher Cieri and
                  Thierry Declerck and
                  Sara Goggi and
                  Hitoshi Isahara and
                  Bente Maegaard and
                  Joseph Mariani and
                  H{\'{e}}l{\`{e}}ne Mazo and
                  Asunci{\'{o}}n Moreno and
                  Jan Odijk and
                  Stelios Piperidis},
  title        = {TableBank: Table Benchmark for Image-based Table Detection and Recognition},
  booktitle    = {Proceedings of The 12th Language Resources and Evaluation Conference,
                  {LREC} 2020, Marseille, France, May 11-16, 2020},
  pages        = {1918--1925},
  publisher    = {European Language Resources Association},
  year         = {2020},
  url          = {https://aclanthology.org/2020.lrec-1.236/},
  bibsource    = {dblp computer science bibliography, https://dblp.org}
}

@inproceedings{DBLP:conf/nips/MoosaviRRG21,
  author       = {Nafise Sadat Moosavi and
                  Andreas R{\"{u}}ckl{\'{e}} and
                  Dan Roth and
                  Iryna Gurevych},
  editor       = {Joaquin Vanschoren and
                  Sai{-}Kit Yeung},
  title        = {SciGen: a Dataset for Reasoning-Aware Text Generation from Scientific
                  Tables},
  booktitle    = {Proceedings of the Neural Information Processing Systems Track on
                  Datasets and Benchmarks 1, NeurIPS Datasets and Benchmarks 2021, December
                  2021, virtual},
  year         = {2021},
  url          = {https://datasets-benchmarks-proceedings.neurips.cc/paper/2021/hash/149e9677a5989fd342ae44213df68868-Abstract-round2.html},
  bibsource    = {dblp computer science bibliography, https://dblp.org}
}

@inproceedings{DBLP:conf/emnlp/LuPLNK23,
  author       = {Xinyuan Lu and
                  Liangming Pan and
                  Qian Liu and
                  Preslav Nakov and
                  Min{-}Yen Kan},
  editor       = {Houda Bouamor and
                  Juan Pino and
                  Kalika Bali},
  title        = {{SCITAB:} {A} Challenging Benchmark for Compositional Reasoning and
                  Claim Verification on Scientific Tables},
  booktitle    = {Proceedings of the 2023 Conference on Empirical Methods in Natural
                  Language Processing, {EMNLP} 2023, Singapore, December 6-10, 2023},
  pages        = {7787--7813},
  publisher    = {Association for Computational Linguistics},
  year         = {2023},
  url          = {https://doi.org/10.18653/v1/2023.emnlp-main.483},
  doi          = {10.18653/V1/2023.EMNLP-MAIN.483},
  bibsource    = {dblp computer science bibliography, https://dblp.org}
}

@inproceedings{DBLP:conf/emnlp/ReimersG19,
  author       = {Nils Reimers and
                  Iryna Gurevych},
  editor       = {Kentaro Inui and
                  Jing Jiang and
                  Vincent Ng and
                  Xiaojun Wan},
  title        = {Sentence-BERT: Sentence Embeddings using Siamese BERT-Networks},
  booktitle    = {Proceedings of the 2019 Conference on Empirical Methods in Natural
                  Language Processing and the 9th International Joint Conference on
                  Natural Language Processing, {EMNLP-IJCNLP} 2019, Hong Kong, China,
                  November 3-7, 2019},
  pages        = {3980--3990},
  publisher    = {Association for Computational Linguistics},
  year         = {2019},
  url          = {https://doi.org/10.18653/v1/D19-1410},
  doi          = {10.18653/V1/D19-1410},
  bibsource    = {dblp computer science bibliography, https://dblp.org}
}

@article{fleiss1971measuring,
  title={Measuring nominal scale agreement among many raters.},
  author={Fleiss, Joseph L},
  journal={Psychological bulletin},
  volume={76},
  number={5},
  pages={378},
  year={1971},
  publisher={American Psychological Association}
}

@inproceedings{DBLP:conf/emnlp/WolfDSCDMCRLFDS20,
  author       = {Thomas Wolf and
                  Lysandre Debut and
                  Victor Sanh and
                  Julien Chaumond and
                  Clement Delangue and
                  Anthony Moi and
                  Pierric Cistac and
                  Tim Rault and
                  R{\'{e}}mi Louf and
                  Morgan Funtowicz and
                  Joe Davison and
                  Sam Shleifer and
                  Patrick von Platen and
                  Clara Ma and
                  Yacine Jernite and
                  Julien Plu and
                  Canwen Xu and
                  Teven Le Scao and
                  Sylvain Gugger and
                  Mariama Drame and
                  Quentin Lhoest and
                  Alexander M. Rush},
  editor       = {Qun Liu and
                  David Schlangen},
  title        = {Transformers: State-of-the-Art Natural Language Processing},
  booktitle    = {Proceedings of the 2020 Conference on Empirical Methods in Natural
                  Language Processing: System Demonstrations, {EMNLP} 2020 - Demos,
                  Online, November 16-20, 2020},
  pages        = {38--45},
  publisher    = {Association for Computational Linguistics},
  year         = {2020},
  url          = {https://doi.org/10.18653/v1/2020.emnlp-demos.6},
  doi          = {10.18653/V1/2020.EMNLP-DEMOS.6},
  bibsource    = {dblp computer science bibliography, https://dblp.org}
}

@inproceedings{DBLP:conf/acl/0001ZL22,
  author       = {Xueqing Wu and
                  Jiacheng Zhang and
                  Hang Li},
  editor       = {Smaranda Muresan and
                  Preslav Nakov and
                  Aline Villavicencio},
  title        = {Text-to-Table: {A} New Way of Information Extraction},
  booktitle    = {Proceedings of the 60th Annual Meeting of the Association for Computational
                  Linguistics (Volume 1: Long Papers), {ACL} 2022, Dublin, Ireland,
                  May 22-27, 2022},
  pages        = {2518--2533},
  publisher    = {Association for Computational Linguistics},
  year         = {2022},
  url          = {https://doi.org/10.18653/v1/2022.acl-long.180},
  doi          = {10.18653/V1/2022.ACL-LONG.180},
  bibsource    = {dblp computer science bibliography, https://dblp.org}
}

@inproceedings{DBLP:conf/acl/LiWSZWS23,
  author       = {Tong Li and
                  Zhihao Wang and
                  Liangying Shao and
                  Xuling Zheng and
                  Xiaoli Wang and
                  Jinsong Su},
  editor       = {Anna Rogers and
                  Jordan L. Boyd{-}Graber and
                  Naoaki Okazaki},
  title        = {A Sequence-to-Sequence{\&}Set Model for Text-to-Table Generation},
  booktitle    = {Findings of the Association for Computational Linguistics: {ACL} 2023,
                  Toronto, Canada, July 9-14, 2023},
  pages        = {5358--5370},
  publisher    = {Association for Computational Linguistics},
  year         = {2023},
  url          = {https://doi.org/10.18653/v1/2023.findings-acl.330},
  doi          = {10.18653/V1/2023.FINDINGS-ACL.330},
  bibsource    = {dblp computer science bibliography, https://dblp.org}
}

@article{DBLP:journals/corr/abs-2403-14457,
  author       = {Anirudh Sundar and
                  Christopher Richardson and
                  Larry Heck},
  title        = {gTBLS: Generating Tables from Text by Conditional Question Answering},
  journal      = {CoRR},
  volume       = {abs/2403.14457},
  year         = {2024},
  url          = {https://doi.org/10.48550/arXiv.2403.14457},
  doi          = {10.48550/ARXIV.2403.14457},
  eprinttype    = {arXiv},
  eprint       = {2403.14457},
  bibsource    = {dblp computer science bibliography, https://dblp.org}
}

@article{DBLP:journals/corr/abs-2309-08963,
  author       = {Xiangru Tang and
                  Yiming Zong and
                  Jason Phang and
                  Yilun Zhao and
                  Wangchunshu Zhou and
                  Arman Cohan and
                  Mark Gerstein},
  title        = {Struc-Bench: Are Large Language Models Really Good at Generating Complex
                  Structured Data?},
  journal      = {CoRR},
  volume       = {abs/2309.08963},
  year         = {2023},
  url          = {https://doi.org/10.48550/arXiv.2309.08963},
  doi          = {10.48550/ARXIV.2309.08963},
  eprinttype    = {arXiv},
  eprint       = {2309.08963},
  bibsource    = {dblp computer science bibliography, https://dblp.org}
}

@article{HeaPA,
  author       = {Weiqi Wang and
                  Xin Liu and
                  Binxuan Huang and
                  Hejie Cui and
                  Rongzhi Zhang and
                  Changlong Yu and
                  Shuowei Jin and
                  Jingfeng Yang and
                  Qingyu Yin and
                  Zhengyang Wang and
                  Zheng Li and
                  Yifan Gao and
                  Priyanka Nigam and
                  Bing Yin and
                  Lihong Li and
                  Yangqiu Song},
  title        = {HeaPA: Difficulty-Aware Heap Sampling and On-Policy Query Augmentation
                  for {LLM} Reinforcement Learning},
  journal      = {CoRR},
  volume       = {abs/2601.22448},
  year         = {2026},
  url          = {https://doi.org/10.48550/arXiv.2601.22448},
  doi          = {10.48550/ARXIV.2601.22448},
  eprinttype   = {arXiv},
  eprint       = {2601.22448},
  bibsource    = {dblp computer science bibliography, https://dblp.org}
}

@inproceedings{DBLP:conf/acl/WangCLNX0SLGLYB25,
  author       = {Weiqi Wang and
                  Limeng Cui and
                  Xin Liu and
                  Sreyashi Nag and
                  Wenju Xu and
                  Chen Luo and
                  Sheikh Muhammad Sarwar and
                  Yang Li and
                  Hansu Gu and
                  Hui Liu and
                  Changlong Yu and
                  Jiaxin Bai and
                  Yifan Gao and
                  Haiyang Zhang and
                  Qi He and
                  Shuiwang Ji and
                  Yangqiu Song},
  editor       = {Wanxiang Che and
                  Joyce Nabende and
                  Ekaterina Shutova and
                  Mohammad Taher Pilehvar},
  title        = {EcomScriptBench: {A} Multi-task Benchmark for E-commerce Script Planning
                  via Step-wise Intention-Driven Product Association},
  booktitle    = {Proceedings of the 63rd Annual Meeting of the Association for Computational
                  Linguistics (Volume 1: Long Papers), {ACL} 2025, Vienna, Austria,
                  July 27 - August 1, 2025},
  pages        = {1--22},
  publisher    = {Association for Computational Linguistics},
  year         = {2025},
  url          = {https://aclanthology.org/2025.acl-long.1/},
  bibsource    = {dblp computer science bibliography, https://dblp.org}
}

@inproceedings{DBLP:conf/acl/0001S25,
  author       = {Weiqi Wang and
                  Yangqiu Song},
  editor       = {Wanxiang Che and
                  Joyce Nabende and
                  Ekaterina Shutova and
                  Mohammad Taher Pilehvar},
  title        = {{MARS:} Benchmarking the Metaphysical Reasoning Abilities of Language
                  Models with a Multi-task Evaluation Dataset},
  booktitle    = {Proceedings of the 63rd Annual Meeting of the Association for Computational
                  Linguistics (Volume 1: Long Papers), {ACL} 2025, Vienna, Austria,
                  July 27 - August 1, 2025},
  pages        = {1568--1596},
  publisher    = {Association for Computational Linguistics},
  year         = {2025},
  url          = {https://aclanthology.org/2025.acl-long.79/},
  bibsource    = {dblp computer science bibliography, https://dblp.org}
}

@inproceedings{DBLP:conf/emnlp/WangFSXDZFBLLS25,
  author       = {Weiqi Wang and
                  Tianqing Fang and
                  Haochen Shi and
                  Baixuan Xu and
                  Wenxuan Ding and
                  Liyu Zhang and
                  Wei Fan and
                  Jiaxin Bai and
                  Haoran Li and
                  Xin Liu and
                  Yangqiu Song},
  editor       = {Christos Christodoulopoulos and
                  Tanmoy Chakraborty and
                  Carolyn Rose and
                  Violet Peng},
  title        = {On the Role of Entity and Event Level Conceptualization in Generalizable
                  Reasoning: {A} Survey of Tasks, Methods, Applications, and Future
                  Directions},
  booktitle    = {Findings of the Association for Computational Linguistics: {EMNLP}
                  2025, Suzhou, China, November 4-9, 2025},
  pages        = {2260--2281},
  publisher    = {Association for Computational Linguistics},
  year         = {2025},
  url          = {https://aclanthology.org/2025.findings-emnlp.122/},
  bibsource    = {dblp computer science bibliography, https://dblp.org}
}

@inproceedings{DBLP:conf/acl/0001FLS0XWBLJCS24,
  author       = {Weiqi Wang and
                  Tianqing Fang and
                  Chunyang Li and
                  Haochen Shi and
                  Wenxuan Ding and
                  Baixuan Xu and
                  Zhaowei Wang and
                  Jiaxin Bai and
                  Xin Liu and
                  Cheng Jiayang and
                  Chunkit Chan and
                  Yangqiu Song},
  editor       = {Lun{-}Wei Ku and
                  Andre Martins and
                  Vivek Srikumar},
  title        = {{CANDLE:} Iterative Conceptualization and Instantiation Distillation
                  from Large Language Models for Commonsense Reasoning},
  booktitle    = {Proceedings of the 62nd Annual Meeting of the Association for Computational
                  Linguistics (Volume 1: Long Papers), {ACL} 2024, Bangkok, Thailand,
                  August 11-16, 2024},
  pages        = {2351--2374},
  publisher    = {Association for Computational Linguistics},
  year         = {2024},
  url          = {https://doi.org/10.18653/v1/2024.acl-long.128},
  doi          = {10.18653/V1/2024.ACL-LONG.128},
  bibsource    = {dblp computer science bibliography, https://dblp.org}
}

@inproceedings{DBLP:conf/emnlp/OuWSJSCJWLMHWMRKD25,
  author       = {Jiefu Ou and
                  William Gantt Walden and
                  Kate Sanders and
                  Zhengping Jiang and
                  Kaiser Sun and
                  Jeffrey Cheng and
                  William Jurayj and
                  Miriam Wanner and
                  Shaobo Liang and
                  Candice Morgan and
                  Seunghoon Han and
                  Weiqi Wang and
                  Chandler May and
                  Hannah Recknor and
                  Daniel Khashabi and
                  Benjamin Van Durme},
  editor       = {Christos Christodoulopoulos and
                  Tanmoy Chakraborty and
                  Carolyn Rose and
                  Violet Peng},
  title        = {{CLAIMCHECK:} How Grounded are {LLM} Critiques of Scientific Papers?},
  booktitle    = {Findings of the Association for Computational Linguistics: {EMNLP}
                  2025, Suzhou, China, November 4-9, 2025},
  pages        = {21712--21735},
  publisher    = {Association for Computational Linguistics},
  year         = {2025},
  url          = {https://aclanthology.org/2025.findings-emnlp.1185/},
  bibsource    = {dblp computer science bibliography, https://dblp.org}
}

@article{DBLP:journals/corr/abs-2504-13834,
  author       = {Muhan Gao and
                  Jash Shah and
                  Weiqi Wang and
                  Daniel Khashabi},
  title        = {Science Hierarchography: Hierarchical Organization of Science Literature},
  journal      = {CoRR},
  volume       = {abs/2504.13834},
  year         = {2025},
  url          = {https://doi.org/10.48550/arXiv.2504.13834},
  doi          = {10.48550/ARXIV.2504.13834},
  eprinttype   = {arXiv},
  eprint       = {2504.13834},
  bibsource    = {dblp computer science bibliography, https://dblp.org}
}

@inproceedings{DBLP:conf/acl/WangFXBSC23,
  author       = {Weiqi Wang and
                  Tianqing Fang and
                  Baixuan Xu and
                  Chun Yi Louis Bo and
                  Yangqiu Song and
                  Lei Chen},
  editor       = {Anna Rogers and
                  Jordan L. Boyd{-}Graber and
                  Naoaki Okazaki},
  title        = {{CAT:} {A} Contextualized Conceptualization and Instantiation Framework
                  for Commonsense Reasoning},
  booktitle    = {Proceedings of the 61st Annual Meeting of the Association for Computational
                  Linguistics (Volume 1: Long Papers), {ACL} 2023, Toronto, Canada,
                  July 9-14, 2023},
  pages        = {13111--13140},
  publisher    = {Association for Computational Linguistics},
  year         = {2023},
  url          = {https://doi.org/10.18653/v1/2023.acl-long.733},
  doi          = {10.18653/V1/2023.ACL-LONG.733},
  bibsource    = {dblp computer science bibliography, https://dblp.org}
}

@inproceedings{DBLP:conf/emnlp/WangF0XLSB23,
  author       = {Weiqi Wang and
                  Tianqing Fang and
                  Wenxuan Ding and
                  Baixuan Xu and
                  Xin Liu and
                  Yangqiu Song and
                  Antoine Bosselut},
  editor       = {Houda Bouamor and
                  Juan Pino and
                  Kalika Bali},
  title        = {{CAR:} Conceptualization-Augmented Reasoner for Zero-Shot Commonsense
                  Question Answering},
  booktitle    = {Findings of the Association for Computational Linguistics: {EMNLP}
                  2023, Singapore, December 6-10, 2023},
  series       = {Findings of {ACL}},
  pages        = {13520--13545},
  publisher    = {Association for Computational Linguistics},
  year         = {2023},
  url          = {https://doi.org/10.18653/v1/2023.findings-emnlp.902},
  doi          = {10.18653/V1/2023.FINDINGS-EMNLP.902},
  bibsource    = {dblp computer science bibliography, https://dblp.org}
}

@article{Qwen3,
  author       = {Qwen Team},
  title        = {Qwen3 Technical Report},
  journal      = {CoRR},
  volume       = {abs/2505.09388},
  year         = {2025},
  url          = {https://doi.org/10.48550/arXiv.2505.09388},
  doi          = {10.48550/ARXIV.2505.09388},
  eprinttype   = {arXiv},
  eprint       = {2505.09388},
  bibsource    = {dblp computer science bibliography, https://dblp.org}
}

\clearpage 

\appendix
\begin{center}
    {\Large\textbf{Appendices}}
\end{center}

\section{Additional Analysis}
\label{appendix:additional_analysis}
\subsection{Method Efficiency Evaluations}
In addition to quality metrics (Table~\ref{tab:maintable_eval_results}), we compare generation efficiency based on (i) generation success rate (GSR) under the backbone context window limit, (ii) average token usage per table, and (iii) average runtime.
These statistics are reported in Table~\ref{tab:cost_analysis}.
Overall, our method achieves a 100\% success rate while keeping token usage controlled, indicating that iterative batching improves robustness without incurring excessive context overhead.

To assess the efficiency and scalability of our iterative batch-based method, we report computational statistics in Table~\ref{tab:cost_analysis}. 
Each method was run using the same LLaMA-3.3 model backend. We measure three aspects: (1) generation success rate, defined as the proportion of prompts yielding complete tables within the context window, (2) average token usage per table, and (3) average runtime per table.
Our method achieves a 100\% success rate, outperforming the baselines that occasionally fail due to context limitations or prompt instability. 
While our runtime is moderately longer than Baseline 1 and Baseline 2, it remains comparable to Newman et al. and stays well within acceptable latency for practical usage. 
Furthermore, token usage remains controlled, confirming that our iterative approach does not incur excessive computational cost despite its multi-step structure. 
These results demonstrate that our method offers a favorable trade-off between performance and efficiency.

\subsection{Additional Backbone Comparisons (GPT-5.1 and GPT-o3)}
\label{app:strong_backbones}
To test whether the relative improvements of our iterative batch-based method persist under stronger proprietary backbones, we run additional comparisons with GPT-5.1 and GPT-o3 under the same evaluation protocol as in the main experiments.
Due to cost constraints, GPT-o3 results are reported on a fixed 50-table subset.
We report T3 utilization metrics (Paper/Schema/Unary/Pairwise F1 and their average) in Table~\ref{tab:strong_backbones}.

\begin{table*}[t]
\small
\centering
\resizebox{!}{!}{
\begin{tabular}{l|ccccc}
\toprule
Backbone \& Method & Paper F1 & Schema F1 & Unary F1 & Pairwise F1 & Avg \\
\midrule
GPT-5.1 + Baseline 2                       & 76.0 & 55.0 & 49.0 & 38.0 & 47.3 \\
GPT-5.1 + Ours                             & 78.2 & 59.3 & 59.3 & 54.5 & 57.8 \\
GPT-o3 + Baseline 2 (50-table subset)      & 75.0 & 53.0 & 47.0 & 36.0 & 45.3 \\
GPT-o3 + Ours (50-table subset / full set) & 77.2 & 58.3 & 58.3 & 53.5 & 56.7 \\
\bottomrule
\end{tabular}
}
\caption{Results under stronger proprietary backbones using the same utilization-based evaluation protocol. GPT-o3 rows are evaluated on a fixed subset due to cost; GPT-5.1 rows follow the same evaluation setup as the corresponding main experiments.}
\label{tab:strong_backbones}
\end{table*}

\subsection{Evaluator Independence Checks}
\label{app:evaluator_checks}
This section tests whether our utilization-based evaluation depends on which LLM synthesizes QA pairs versus which LLM answers them. 
We decouple \emph{QA synthesis} and \emph{answering}: a synthesizer produces all schema/unary/pairwise QAs from a table; a distinct answerer then answers those QAs from the comparison table. 
We then swap roles and recompute scores. Stability is measured by (i) rank correlations of method orderings (Spearman, Kendall) over model$\times$method tuples, and (ii) the maximum absolute F1 change per dimension (Schema/Unary/Pairwise).

\begin{table}[t]
\small
\centering
\begin{tabular}{l|cc}
    \toprule
    Method & GSR & \#Tokens \\
    \midrule
    Baseline 1 & 48.19\% & 128K \\
    Baseline 2 & 98.23\% & 167K \\
    Newman et al. & 99.71\% & 110K \\
    \textbf{Ours} & 100.0\% & 118K \\
    \bottomrule
    \end{tabular}
\caption{Comparison of the efficiency of different methods. GSR stands for generation success rate.}
\label{tab:method_efficiency}
\end{table}

\begin{table}[t]
\small
\centering
\resizebox{\linewidth}{!}{
\begin{tabular}{l|ccc}
    \toprule
    Statistic & Paper Count & Column Count & Distractor Count \\
    \midrule
    Min   & 1    & 2    & 4  \\
    Max   & 32   & 13   & 10 \\
    Mean  & 3.65 & 3.56 & 5.21 \\
    Total & 7158 & 6967 & 10196 \\
    \bottomrule
\end{tabular}
}
\caption{Summary statistics of the \dataset{} benchmark. We report aggregate values for the number of papers, columns, and distractor papers per table.}
\label{tab:dataset_stats}
\end{table}

\begin{figure}[t]
     \centering
     \includegraphics[width=1\linewidth]{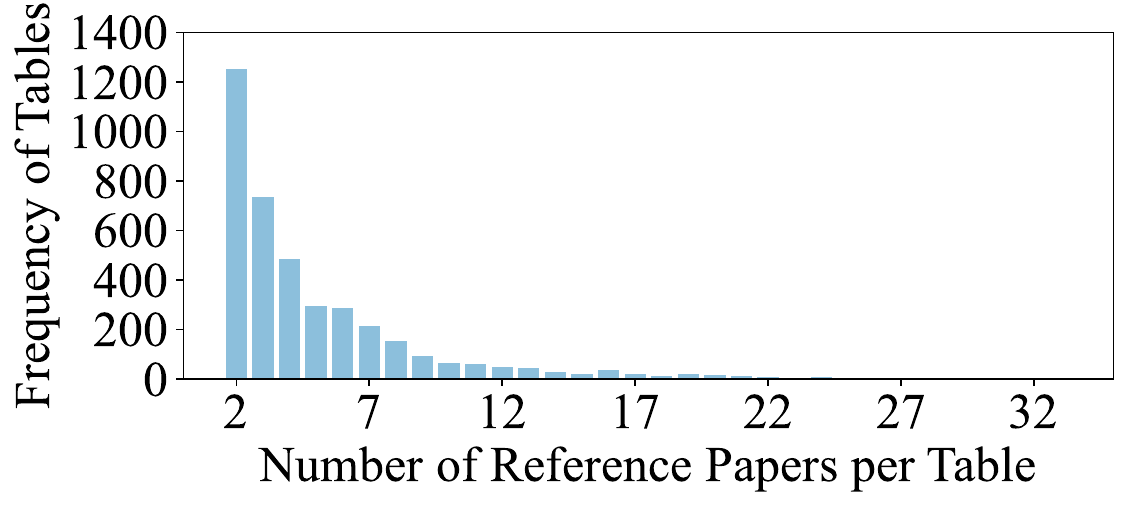}
     \caption{Distribution of number of papers in each table.}
    \label{fig:number_papers_distribution}
\end{figure}

\begin{table}[t]
\small
\centering
\resizebox{\linewidth}{!}{
\begin{tabular}{@{}l|ccc@{}}
    \toprule
    Method & Success Rate & \#Tokens & Avg. Runtime \\
    \midrule
    Baseline 1 & 48.2\% & 128K & 37 \\
    Baseline 2 & 98.2\% & 167K & 118 \\
    Newman et al. & 99.7\% & 110K & 208 \\
    Ours     & 100.0\% & 118K & 194 \\
    \bottomrule
\end{tabular}
}
\caption{Computational cost and efficiency metrics across different generation methods using LLaMA-3.3. We report the generation success rate, average token usage, and average runtime (s) per table.}
\label{tab:cost_analysis}
\end{table}

\textbf{Setup.} We evaluate four synthesizer$\to$answerer pairs: GPT-4o$\to$GPT-4o-mini, GPT-4o-mini$\to$GPT-4o, GPT-4o$\to$Llama-3.3, Llama-3.3$\to$GPT-4o (Table~\ref{tab:evaluator_rank}), and additionally include a cross-family swap using \texttt{Qwen3-30B-A3B-Instruct} (Table~\ref{tab:qwen3_evaluator_swap}).
We randomly sample 50 tables (stratified by table size) and include all generation methods (Baseline~1, Baseline~2, Newman et al., Ours). 
Decoding temperature is 0.5; the same seeds as the main runs are reused. Each run uses the same QA quantity as the main evaluation (all schema and unary QAs, 10 pairwise QAs per table).

\begin{table*}[h]
\small\centering
\begin{tabular}{@{}l|cc|cc|cc@{}}
\toprule
\multirow{2}{*}{Synth$\to$Ans} & \multicolumn{2}{c|}{Schema F1} & \multicolumn{2}{c|}{Unary F1} & \multicolumn{2}{c}{Pairwise F1} \\
& Mean $\Delta$ & Max $\Delta$ & Mean $\Delta$ & Max $\Delta$ & Mean $\Delta$ & Max $\Delta$ \\
\midrule
4o $\to$ 4o-mini       & 0.4 & 1.5 & 0.5 & 1.6 & 0.6 & 1.8 \\
4o-mini $\to$ 4o       & 0.5 & 1.7 & 0.6 & 1.8 & 0.7 & 1.9 \\
4o $\to$ Llama3.3      & 0.6 & 1.9 & 0.7 & 2.0 & 0.9 & 2.1 \\
Llama3.3 $\to$ 4o      & 0.5 & 1.7 & 0.6 & 1.9 & 0.8 & 2.0 \\
\bottomrule
\end{tabular}
\caption{Absolute F1 drift (percentage points) when swapping the QA synthesizer and answerer. Values are averaged over model$\times$method tuples.}
\label{tab:evaluator_drift}
\end{table*}

\begin{table*}[h]
\small\centering
\resizebox{\linewidth}{!}{
\begin{tabular}{@{}l|cc|cc@{}}
\toprule
Synth$\to$Ans & Spearman (Schema/Unary/Pairwise) $\uparrow$ & Kendall (Schema/Unary/Pairwise) $\uparrow$ & Macro Spearman $\uparrow$ & Macro Kendall $\uparrow$ \\
\midrule
4o $\to$ 4o-mini  & 0.87 / 0.88 / 0.86 & 0.71 / 0.72 / 0.70 & 0.87 & 0.71 \\
4o-mini $\to$ 4o  & 0.85 / 0.86 / 0.86 & 0.69 / 0.70 / 0.70 & 0.86 & 0.70 \\
4o $\to$ Llama3.3 & 0.83 / 0.84 / 0.85 & 0.67 / 0.68 / 0.69 & 0.84 & 0.68 \\
Llama3.3 $\to$ 4o & 0.84 / 0.85 / 0.86 & 0.68 / 0.69 / 0.70 & 0.85 & 0.69 \\
\bottomrule
\end{tabular}
}
\caption{Ranking stability across methods under evaluator swaps. Correlations computed over model$\times$method tuples.}
\label{tab:evaluator_rank}
\end{table*}

\textbf{Result.} Across swaps, macro Kendall $\approx 0.68$–$0.71$ and macro Spearman $\approx 0.84$–$0.87$, while max absolute F1 changes remain $\leq 2.1$ points. Method rankings are therefore stable, and absolute performance differences are small, suggesting limited evaluator dependence and addressing circularity concerns.

\paragraph{Cross-family evaluator swap with Qwen3.}
In addition to the swaps above, we also test evaluator independence by replacing GPT-4o with the open-source \texttt{Qwen3-30B-A3B-Instruct}~\cite{Qwen3} as either the QA synthesizer or the answerer on a subset.
We report the relative change in macro-F1 (vs.\ GPT-4o as the default evaluator) and ranking stability in Table~\ref{tab:qwen3_evaluator_swap}.

\begin{table*}[t]
\small
\centering
\begin{tabular}{l|ccc}
\toprule
Synthesizer \& Answerer & Relative $\Delta$F1 vs.\ GPT-4o & Spearman $\rho$ & Kendall $\tau$ \\
\midrule
GPT-4o $\to$ Qwen3-30B-A3B & +1.7 & 0.89 & 0.78 \\
Qwen3-30B-A3B $\to$ GPT-4o & +2.4 & 0.86 & 0.74 \\
\bottomrule
\end{tabular}
\caption{Cross-family evaluator swap using \texttt{Qwen3-30B-A3B-Instruct}. Relative $\Delta$F1 is computed against the default GPT-4o-based evaluation on the same subset; rank correlations are computed over model$\times$method tuples.}
\label{tab:qwen3_evaluator_swap}
\end{table*}

Together with Tables~\ref{tab:evaluator_drift}--\ref{tab:evaluator_rank}, these results indicate that method rankings are stable under evaluator changes across both proprietary and open-source model families.

\begin{table}[t]
\small
\centering
\begin{tabular}{p{0.18\linewidth}|p{0.74\linewidth}}
\toprule
\textbf{Constraint} & \textbf{Operationalization (what we enforce)} \\
\midrule
Self-contained & Mentions the topic and the intended comparison goal without assuming access to the gold table. \\
Goal-oriented & Specifies what decision/analysis the table should support (e.g., comparing families of methods, tracing trends, or summarizing benchmark settings). \\
Non-leaking & Avoids verbatim column headers and specific numeric/string values from the gold table; avoids directly paraphrasing headers. \\
Concise & 1--2 sentences; avoids long enumerations of fields to keep the request natural. \\
\bottomrule
\end{tabular}
\caption{Checklist used for caption-to-demand rewriting. The automatic leak check (Appendix~\ref{appendix:implementation_details}) flags violations and triggers rewriting.}
\label{tab:user_demand_checklist}
\end{table}

\subsection{Additional Retrieval Baselines and Rationale}
\label{app:retrieval_baselines}
We compare dense and sparse retrieval for constructing candidate pools used in distractor verification. Each table’s user demand is matched against all paper titles and abstracts.
We evaluate TF--IDF, BM25, and a dense retriever based on \texttt{sentence-t5-xxl} (SentenceTransformers) on a 200-table slice.
All methods return top-$k$ candidates; we report macro Precision@$k$ (P@$k$), Recall@$k$ (R@$k$), and Average Precision (AP).
Sparse methods use standard tokenization with stopword removal; dense retrieval encodes the concatenation of title and abstract and ranks by cosine similarity.

\begin{table}[h]
\small\centering
\begin{tabular}{@{}l|ccc@{}}
\toprule
Method & P@10 & R@10 & AP \\
\midrule
TF--IDF       & 0.44 & 0.36 & 0.31 \\
BM25          & 0.49 & 0.40 & 0.36 \\
SentenceBERT  & \textbf{0.54} & \textbf{0.47} & \textbf{0.42} \\
\bottomrule
\end{tabular}
\caption{Top-10 retrieval quality (macro). Dense retrieval improves recall and AP, which is critical for minimizing false negatives in later filtering.}
\label{tab:retrieval_top10}
\end{table}

We additionally probe different cutoffs to assess stability. Results at smaller and larger $k$ show the same ordering, with the dense method maintaining a recall advantage as $k$ increases.

\begin{table}[h]
\small\centering
\begin{tabular}{@{}l|cc|cc@{}}
\toprule
& \multicolumn{2}{c|}{@5} & \multicolumn{2}{c}{@20} \\
Method & P@5 & R@5 & P@20 & R@20 \\
\midrule
TF--IDF       & 0.58 & 0.25 & 0.35 & 0.49 \\
BM25          & 0.62 & 0.28 & 0.38 & 0.55 \\
SentenceBERT  & \textbf{0.65} & \textbf{0.33} & \textbf{0.40} & \textbf{0.62} \\
\bottomrule
\end{tabular}
\caption{Precision/recall at @5 and @20. Dense retrieval sustains higher recall across cutoffs.}
\label{tab:retrieval_cutoffs}
\end{table}

These results justify our use of an embedding-based retriever to construct a semantically challenging yet realistic candidate pool. 
Sparse baselines remain competitive in precision at very small cutoffs (e.g., @5), but they under-recall paraphrastic matches common in scientific text. 
The dense approach reduces downstream risk of missing truly relevant papers during human verification.

\section{Implementation Details}
\label{appendix:implementation_details}
In this section, we provide additional implementation details about our benchmark curation and evaluation pipeline, including the prompt we used and the models we accessed.

\subsection{Prompts Used}
We provide all prompts used in our benchmark construction and evaluation. In benchmark construction, the key steps are (i) rewriting captions into user demands without leaking schema/values and (ii) synthesizing QA pairs from the ground-truth table. In evaluation, we normalize tables, canonicalize headers, decouple QA synthesis from answering, and run a reverse-QA pass for precision.

\medskip
\noindent\textbf{User demand rewrite (benchmark construction).}
This prompt rewrites terse/arXiv-style captions into contextually self-contained user demands that specify the table’s purpose without leaking column names or specific values. It improves realism by emulating well-specified but abstract requests while keeping evaluation fair.

\begin{displayquote}
\textbf{\texttt{\small 
Given a literature review table, along with its caption, you are tasked with writing a user demand or intention for the creator of this table. The user demand should be written as though you are instructing an AI system to generate the table. Avoid directly mentioning column names in the table itself, but instead, focus on explaining why the table is needed and what information it should contain. You may include a description of the table’s structure, whether it requires detailed or summarized columns. Additionally, infer the user's intentions from the titles of the papers the table will include. Limit each user demand to 1-2 sentences.
Examples of good user demands are:
I need a table that outlines how each study conceptualizes the problem, categorizes the task, describes the data analyzed, and summarizes the main findings. The table should have detailed columns for each of these aspects.
Generate a detailed table comparing the theoretical background, research methodology, and key results of these papers. You can use several columns to capture these aspects for each paper.
I want to create a table that summarizes the datasets used to evaluate different GNN models, focusing on the common features and characteristics found across the papers listed below. The table should have concise columns to highlight these dataset attributes.
Now, write a user demand for the table below. The caption of the table is \textcolor{headcolor}{``\texttt{<CAPTION>}''}. The table looks like this:\\
\textcolor{headcolor}{\texttt{<TABLE>}}\\
The following papers are included in the table:\\
\textcolor{headcolor}{\texttt{<PAPER-1>}~\ldots~\texttt{<PAPER-N>}}\\
Write the user demand for this table. Do not include the column names in the user demand. Write a concise and clear user demand covering the function, topic, and structure of the table with one or two sentences. The user demand is:
}}
\end{displayquote}

\medskip
\noindent\textbf{Leak check for user demands (guardrail).}
Immediately after rewriting, we run a leak check to ensure the demand does not expose schema labels or table values, and to auto-rewrite if needed. This keeps construction oracle-free and reproducible.

\begin{displayquote}
\textbf{\texttt{\small
You are given: (i) a user demand written from a caption, and (ii) the target table (schema and a few cell values). \\
Decide if the user demand \emph{leaks} any column names or specific cell values, or directly paraphrases them. \\
Return \{ACCEPT, REWRITE\} and a one-sentence reason. If REWRITE, produce a version that removes leaked tokens while staying specific and self-contained. \\
Output JSON only: \{"decision": "...", "reason": "...", "demand": "..." \}
}}
\end{displayquote}

\medskip
\noindent\textbf{QA synthesis from ground truth (recall side).}
This prompt generates binary QA pairs from the gold table to test whether a system table retains schema, values, and pairwise relations. It provides full coverage for schema/unary and a sampled, diverse set for pairwise.

\begin{displayquote}
\textbf{\texttt{\small 
You will evaluate the quality of a generated table by comparing it against a ground-truth table. The goal is to assess whether the generated table correctly retains the schema, individual values, and pairwise relationships. This is achieved by generating targeted QA pairs based on the ground-truth table and answering them using the generated table.
Step 1: QA Pair Generation Based on the Ground-Truth Table
Generate binary (Yes/No) QA pairs focusing on three aspects:
Schema QA Pairs: Check whether a specific column from the ground-truth table appears in the generated table schema.
Example: Is Dataset included in the table schema?
Unary Value QA Pairs: Check whether a specific cell value from the ground-truth table is present in the generated table.
Example: Is CL, TL the loss function for paper CN-LexNet?
Pairwise Value QA Pairs: Check whether a relationship between two values remains consistent in the generated table.
Example: Is ResNet-v2 using more evaluation metrics than GAN?
For Schema and Unary Value, generate a QA pair for every column and every cell, respectively. For Pairwise Value, randomly sample 10 pairs per table and construct the corresponding QA pairs.
Step 2: Answering QA Pairs Using the Generated Table
After generating the QA pairs, answer them using the generated table. Provide only "yes" or "no" responses:
If the information is present in the generated table, respond with "yes."
If the information is missing or different, respond with "no."
Your task is to generate the QA pairs based on the ground-truth table and then answer them based on the generated table.
Now, begin by generating the QA pairs.
}}
\end{displayquote}

\medskip
\noindent\textbf{Answering gold QAs on the system table (decoupled answerer).}
When we decouple synthesis from answering (for evaluator-independence checks and robustness), we use an answer-only prompt that strictly binds answers to the system table content.

\begin{displayquote}
\textbf{\texttt{\small
Input: (i) a normalized \emph{generated} table, and (ii) a list of binary Yes/No QAs synthesized from the \emph{ground-truth} table. \\
Answer each QA \emph{using only the generated table}. If explicitly supported and consistent, answer "yes"; otherwise "no". \\
Return one lowercase token per line: yes/no (no explanations).
}}
\end{displayquote}

\medskip
\noindent\textbf{Reverse-QA synthesis from the system table (precision side).}
To measure precision, we synthesize QAs from the system table and answer them on the gold table. This credits only content supported by the gold table and rejects unsupported “novel” entries.

\begin{displayquote}
\textbf{\texttt{\small
Given a normalized \emph{generated} table, create binary Yes/No QAs for: (i) schema presence, (ii) specific cell values, and (iii) pairwise relations (10 pairs). \\
Write unambiguous questions that can be answered solely from this generated table. Avoid trivial or duplicate QAs. \\
Return JSON list: [{"type":"schema|unary|pairwise","q":"..."} ...]
}}
\end{displayquote}

\medskip
\noindent\textbf{Answering reverse-QAs on the gold table (precision side).}
This answer-only prompt binds answers to the gold table. By instruction, any system-only content is answered “no”.

\begin{displayquote}
\textbf{\texttt{\small
Input: (i) a normalized \emph{gold} table, and (ii) a list of Yes/No QAs synthesized from the \emph{generated} table. \\
Answer each QA using only the gold table. Novel content not present in gold must be answered "no". \\
Return one lowercase token per line: yes/no.
}}
\end{displayquote}

\medskip
\noindent\textbf{QA validation and deduplication (quality filter).}
Before answering, we filter QAs to remove ill-formed, non-binary, trivial, or duplicate items. This reduces evaluator brittleness and ensures consistent scoring.

\begin{displayquote}
\textbf{\texttt{\small
Given a list of binary QAs, remove any that are ill-formed, non-binary, trivially true/false, or duplicates. \\
Return the filtered list as JSON in the same schema and include a "removed" list with reasons.
}}
\end{displayquote}

\medskip
\noindent\textbf{Table normalization (robust parsing).}
We normalize both gold and system tables to a rectangular CSV with consistent headers and “N/A” for missing entries so that downstream prompts and scripts operate deterministically.

\begin{displayquote}
\textbf{\texttt{\small
You are given a table in arbitrary Markdown/LaTeX/CSV. Normalize it into a rectangular CSV with a header row, one row per paper. \\
-- Trim whitespace; collapse multi-line cells; preserve units and text verbatim; fill missing cells with "N/A". \\
-- Do not infer new columns or values. \\
Return CSV only (no commentary).
}}
\end{displayquote}

\medskip
\noindent\textbf{Schema alias canonicalization (header mapping).}
We canonicalize header synonyms (e.g., “Dataset” vs “Data”) before computing schema scores to reduce false mismatches due to phrasing.

\begin{displayquote}
\textbf{\texttt{\small
Given two header lists (gold vs system), produce a one-to-one or one-to-many mapping of system headers to gold headers \emph{when they are semantically equivalent}. \\
Rules: prefer exact matches; allow synonyms; keep units consistent; do not map if semantics differ. \\
Return JSON: [{"system": "...", "gold": "...", "justification": "..."} ...]
}}
\end{displayquote}

The distribution of number of papers per table in~\dataset{} is shown in Figure~\ref{fig:number_papers_distribution}.

\subsection{Evaluation Implementations}
We access all open-source LLMs via the Hugging Face library~\cite{DBLP:conf/emnlp/WolfDSCDMCRLFDS20}.
The models used are \texttt{meta-llama/Llama-3.3-70B-Instruct}, \texttt{mistralai/Mistral-Large-Instruct-2411}, and \texttt{deepseek-ai/DeepSeek-V3}.
For GPT models, we access them via the official OpenAI Batch API\footnote{\href{https://platform.openai.com/docs/guides/batch}{https://platform.openai.com/docs/guides/batch}}.
The models used are \texttt{gpt-4o-mini-2024-07-18} and \texttt{gpt-4o-2024-08-06}.
Note that the DeepSeek model family has a context window limit of 64K tokens, whereas the others have a limit of 128K tokens.
The generation temperature is set to 0.5 for all experiments.
All experiments are repeated twice and the average performance is reported.

We normalize both gold and generated tables to CSV grids with a single header row and “N/A” for missing cells, then apply schema-alias canonicalization before scoring. For the recall side, we synthesize QAs from the gold table and answer them on the system table; for the precision side, we synthesize QAs from the system table and answer them on the gold table. Answers are strictly “yes”/“no”; we lowercase, strip punctuation, and parse the first token if extra text is emitted. Schema and unary QAs are exhaustive; for pairwise, we sample 10 pairs per table while avoiding duplicates and promoting coverage over distinct columns. Randomization uses fixed seeds when supported; otherwise we repeat evaluation twice and average.

For cross-evaluator checks, we decouple QA synthesis and answering and swap the evaluator models as reported in Appendix~\ref{app:evaluator_checks}. Decoding parameters use temperature 0.5, top\_p 1.0, no repetition penalties, and a max\_new\_tokens budget that comfortably covers the QA list; timeouts are 120s per call with up to two retries on transient errors. Significance testing for main-table comparisons uses table-level bootstrap (10{,}000 resamples) with 95\% confidence intervals; full CIs and prompt JSON I/O schemas will be released with code.
For the additional cross-family evaluator swap, we also use \texttt{Qwen3-30B-A3B-Instruct} under the same decoding settings~\cite{HeaPA,DBLP:conf/acl/0001FLS0XWBLJCS24,DBLP:conf/emnlp/WangFSXDZFBLLS25,DBLP:conf/acl/WangFXBSC23,DBLP:conf/emnlp/WangF0XLSB23,DBLP:conf/acl/WangCLNX0SLGLYB25,DBLP:conf/acl/0001S25}.

\section{Annotation Details}
\label{appendix:annotation_details}
\begin{figure*}[ht]
     \centering
     \includegraphics[width=0.85\linewidth]{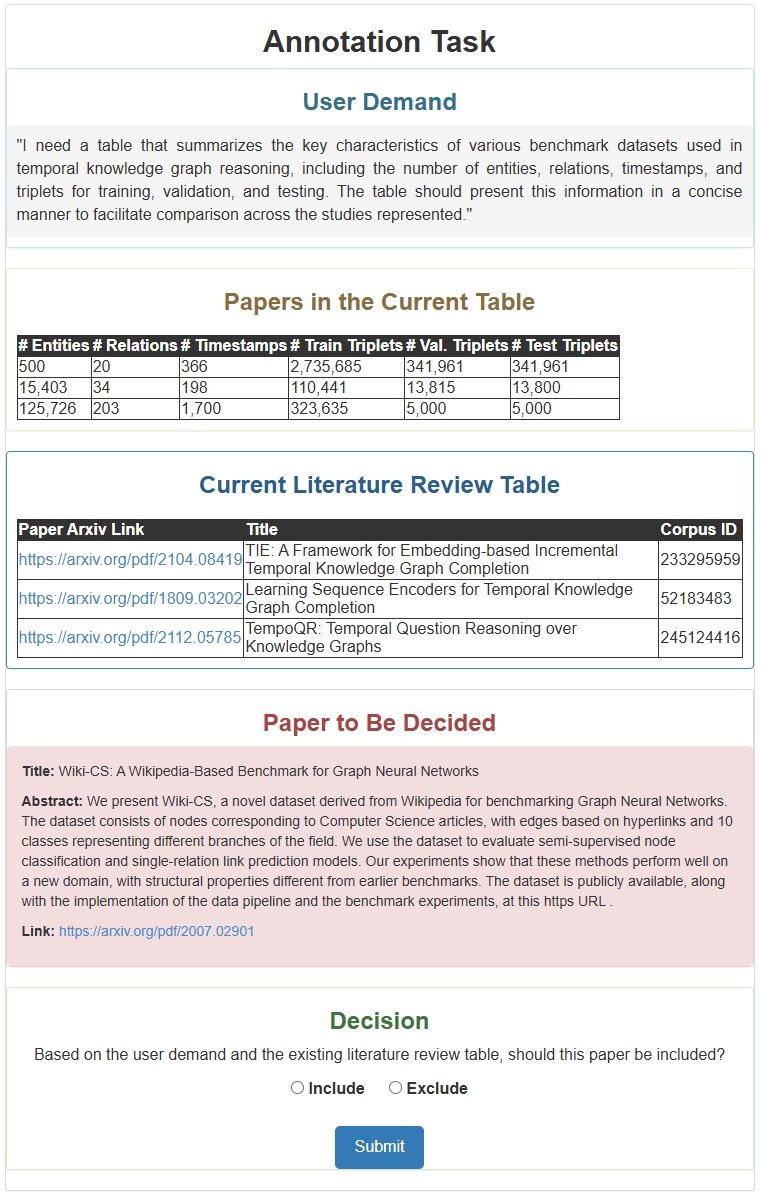}
     \caption{The annotation interface we used for collecting the gold labels for distractor papers.}
    \label{fig:annotation_interface}
\end{figure*}

To ensure high-quality annotations, we recruited 19 trained graduate-level annotators with computer-science research experience and admitted them only after passing qualification rounds. 
For each candidate paper, annotators received clear, layman-friendly instructions with definitions and multiple examples, then confirmed they had read them via a checkbox before starting. 
Using the interface in Figure~\ref{fig:annotation_interface}, each instance was judged with a binary \{include, exclude\} decision relative to the user demand and the known reference papers (title+abstract basis). 
Every instance received two independent labels; on disagreement, two additional adjudicators resolved to consensus. We continuously monitored performance, provided targeted feedback on common errors, and removed spammers or underperformers. 
The study ran for eleven days; first-pass agreement reached 94\% pairwise with Fleiss’ $\kappa{=}0.73$, the self-reported median time per decision was about seven minutes, and annotators were compensated above the local minimum wage. 
The final corpus contains 10{,}196 curated distractor labels across 1{,}957 tables (with 10 initial candidates per table before filtering), supporting the quality reported in Section~\ref{sec:difficulty_noisy_paper_retrieval}.


\section{Case Studies}
\label{appendix:case_studies}
Table~\ref{appendix_tab:user_demand_examples} shows examples of original captions and rewritten user demands, illustrating how short or context-dependent captions can be transformed into self-contained requests that better specify the intended comparison goal.
Table~\ref{appendix_tab:synthesized_QAs} provides illustrative examples of schema, unary, and pairwise questions used by our utilization-based evaluation to measure schema coverage, factual retention, and relational consistency.

We additionally include two example pairs of ground-truth and generated tables in Table~\ref{appendix_tab:case_studies_table}.
These examples highlight common behaviors in literature-review table generation: systems can enrich tables by adding helpful organizing fields, but may also omit important attributes from the ground truth or introduce overly fine-grained fields that dilute the central comparison.
Overall, the examples underscore the importance of jointly handling paper selection, schema induction, and value grounding.

\begin{table*}[t]
  \centering
  \small
  \begin{tabular}{@{}p{0.30\textwidth} | p{0.68\textwidth}@{}}
    \toprule
    \textbf{Original Table Caption} & \textbf{User Demand} \\
    \midrule
    Comparison of trajectory and path planning approaches
    &
    I need a table that compares recent trajectory/path planning methods, emphasizing how they handle safety constraints (e.g., collision avoidance) and what assumptions they make about the environment. Please summarize the key strengths, limitations, and typical application settings for each approach. \\
    \midrule
    Publications with deep-learning focused sampling methods. We cluster the papers based on the space the sample through and how the samples are evaluated. Some approaches further consider an optional refinement stage.
    &
    Please create a table that organizes learning-based sampling methods by how candidates are generated and how they are scored, so it is easy to see the main design choices across papers. If a method uses an extra refinement step or additional supervision, highlight that in the comparison. \\
    \midrule
    Categorization of textual explanation methods.
    &
    I want a table that groups approaches for producing textual explanations by what kind of explanation they generate and what evidence they rely on (e.g., rationales, templates, retrieved facts). The goal is to quickly compare how different families of methods justify their predictions. \\
    \midrule
    Metadata of the three benchmarks that we focus on. XSumSota is a combined benchmark of {{cite:1400aac}} and {{cite:d420ef8}} for summaries generated by the state-of-the-art summarization models.
    &
    Please produce a table that contrasts these benchmarks in terms of what they measure and how they are evaluated, including how labels are obtained and what the evaluation protocol looks like. I want to understand which benchmark is most suitable for comparing modern summarization systems. \\
    \midrule
    Review of open access ground-based forest datasets
    &
    I need a table summarizing open-access forest datasets, focusing on what is recorded, when and where data are collected, and what tasks each dataset supports. The table should make it easy to choose a dataset for a specific forest monitoring objective. \\
    \midrule
    Comparison of existing consistency-type models.
    &
    Please construct a table comparing consistency-oriented models by what notion of consistency they enforce and how they operationalize it (objective, constraints, or training signal). The table should make clear how each model differs in assumptions and what settings it is intended for. \\
    \bottomrule
  \end{tabular}
  \caption{Examples of original captions and rewritten user demands. User demands are written to be self-contained and goal-oriented while avoiding direct leakage of gold headers or specific cell values.}
  \label{appendix_tab:user_demand_examples}
\end{table*}

\begin{table*}[t]
  \centering
  \small
  \begin{tabular}{@{}p{0.33\textwidth} | p{0.33\textwidth} | p{0.33\textwidth}@{}}
    \toprule
    \textbf{Schema QA (presence of a field)} & \textbf{Unary QA (a specific entry)} & \textbf{Pairwise QA (a relation)} \\
    \midrule
    Is the evaluation benchmark included in the table schema?
    &
    Is \texttt{ZsRE} listed as an evaluation benchmark for \texttt{ROME}?
    &
    Does \texttt{ROME} achieve higher reported accuracy than \texttt{MEND} on \texttt{ZsRE}? \\
    \midrule
    Is the training signal/objective described in the table schema?
    &
    Is \texttt{consistency regularization} listed as a training signal for at least one method in the table?
    &
    Do methods that use \texttt{consistency regularization} report better performance than those that do not on the same benchmark? \\
    \midrule
    Is the data source or dataset type included in the table schema?
    &
    Is \texttt{ImageNet} listed as a dataset used in any compared method?
    &
    Is \texttt{Method A} evaluated on more datasets than \texttt{Method B} in the table? \\
    \midrule
    Is the compute setting (e.g., hardware or budget) included in the table schema?
    &
    Is \texttt{8$\times$A100} listed as the hardware setting for any method?
    &
    Does \texttt{Method A} report a lower compute budget than \texttt{Method B} under the same evaluation setup? \\
    \midrule
    Is the publication year or timeframe included in the table schema?
    &
    Is \texttt{2023} listed as the publication year for \texttt{Method X}?
    &
    Are post-2022 methods reported as outperforming pre-2020 methods on the same metric in the table? \\
    \bottomrule
  \end{tabular}
  \caption{Illustrative examples of schema/unary/pairwise questions. In the benchmark, questions are synthesized from each ground-truth table and are answerable by reading the corresponding comparison table.}
  \label{appendix_tab:synthesized_QAs}
\end{table*}

\begin{table*}[]
    \small
    \centering
    \begin{subtable}{\textwidth}
        \centering
        \begin{tabular}{l | c | c}
            \toprule
            Tasks & \#Categories & Evaluation Metric \\
            \midrule
            fine-grained & 100 & mean accuracy \\
            face & 9131 & - \\
            \bottomrule
        \end{tabular}
        \caption{Ground-truth table of the first pair of example.}
    \end{subtable}
    
    \vspace{1em}

    \begin{subtable}{\textwidth}
        \centering
        \begin{tabular}{c | p{2.5cm} | p{3cm} | p{2.5cm} | c}
            \toprule
            Number of Images & Number of Subjects & Avg. Images per Subject & Number of Classes & Dataset Purpose \\
            \midrule
            10,000 & 100 & 100 & 100 & Fine-grained visual classification \\
            3,310,000 & 9,131 & 362.6 & 9,131 & Face recognition across variations \\
            \bottomrule
        \end{tabular}
        \caption{Generated table of the first pair of example.}
    \end{subtable}

    \vspace{1em}

    \begin{subtable}{\textwidth}
        \centering
        \begin{tabular}{l | p{10cm}}
            \toprule
            Problem & Description \\
            \midrule
            Visual Reference Resolution & Capturing related visual region through an associative attention memory. \\
            Visual Reference Resolution & Selectively referring dialogue history to refine the visual attention until referencing the answer. \\
            Visual Reference Resolution & Establishing mapping of visual object and textual entities to exclude undesired visual content. \\
            Visual-based Dialogue Strategies Optimization & Enhancing response generator with discriminator by RL reward. \\
            Visual-based Dialogue Strategies Optimization & Maximizing the information gain while asking questions with a RL paradigm for explicit dialogue goals. \\
            Pre-trained Vision Language Model-based VAD & Training unified Transformer encoder initialized by BERT with two visual training objectives. \\
            Pre-trained Vision Language Model-based VAD & Utilizing GPT-2 to capture cross-modal semantic dependencies. \\
            Unique Training Schemes-based VAD & Simulating Dual-coding theory of human cognition to adaptively find query-related information from the image. \\
            Unique Training Schemes-based VAD & Asking questions to confirm the conjecture of models about the referent guided by human cognitive literature. \\
            \bottomrule
        \end{tabular}
        \caption{Ground-truth table of the second pair of example.}
    \end{subtable}
    
    \vspace{1em}

    \begin{subtable}{\textwidth}
        \centering
        \begin{tabular}{c | p{3cm} | p{2cm} | p{2cm} | p{2cm} | p{2cm} | p{1.5cm}}
            \toprule
            ID & Method Used & Dataset & Problem Addressed & Performance Metric & Results Achieved & Model Type \\
            \midrule
            5677543 & Attention memory model & VisDial & Visual dialog with reference resolution & Answer prediction accuracy & ~16\% improvement over state-of-the-art & Generative \\
            54446647 & Recursive Visual Attention mechanism & VisDial v0.9 & Visual co-reference resolution & Mean Rank & State-of-the-art performance & Generative \\
            236478107 & Multimodal transformer with visual grounding & VisDial v0.9 and v1.0 & Visual dialogue generation & BLEU & Achieves new state-of-the-art results & Generative \\
            24537813 & Adversarial learning with co-attention & VisDial & Visual dialog generation & Recall@5 & +2.14\% improvement over the previous best & Generative \\
            196180698 & Goal-oriented question generation model & GuessWhat?! & Goal-oriented visual dialogue & Accuracy & 67.19\% on GuessWhat?! & Generative \\
            216562638 & Vision-dialog transformer architecture & VisDial v0.9 and v1.0 & Visual dialog & NDCG & New state-of-the-art performance & Generative and Discriminative \\
            220045105 & GPT-2 based architecture & AVSD & Video-grounded dialogue & BLEU & Outperforms existing approaches & Generative \\
            208138178 & Adaptive dual encoding framework & VisDial & Visual dialogue & Accuracy & State-of-the-art results & Generative \\
            237491596 & Beam search re-ranking algorithm & GuessWhat?! & Referential guessing games & Task accuracy & +4.35\% improvement with re-ranking & Generative \\
            \bottomrule
        \end{tabular}
        \caption{Generated table of the second pair of example.}
    \end{subtable}
    \caption{Case studies on the generation of literature review tables in \dataset.}
    \label{appendix_tab:case_studies_table}
\end{table*}

\end{document}